\documentclass[11pt,a4paper]{article}
\usepackage[hyperref]{acl2020}
\usepackage{times}
\usepackage{latexsym}

\aclfinalcopy 
\usepackage{multirow}
\usepackage{amsmath}
\usepackage[font={small}]{caption}
\usepackage{graphicx}
\usepackage{booktabs}
\usepackage{subfig}
 \usepackage{amssymb}
 \usepackage{amsmath}
\usepackage{bbm}
\usepackage{hyperref}
\usepackage{colortbl}
\usepackage{algorithm,algorithmicx,algpseudocode}
\let\OldStatex\Statex
\renewcommand{\Statex}[1][3]{%
  \setlength\@tempdima{\algorithmicindent}%
  \OldStatex\hskip\dimexpr#1\@tempdima\relax}
\definecolor{mygreen}{rgb}{0.0, 0.5, 0.0}
\definecolor{myorange}{rgb}{1.0, 0.49, 0.0}	

\definecolor{mygrey}{rgb}{0.86, 0.86, 0.86}

\title{Towards More Fine-grained and Reliable NLP Performance Prediction}
\date{\today}

\author{Zihuiwen Ye, \quad Pengfei Liu, \quad Jinlan Fu$\dag$, \quad Graham Neubig \\
   Carnegie Mellon University,  $\dag$ Fudan University \\
   \texttt{\{zihuiwey,pliu3,gneubig\}@andrew.cmu.edu},\\  \texttt{fujl16@fudan.edu.cn}}

\begin{document}
\maketitle

\begin{abstract}
    Performance prediction, the task of estimating a system's performance without performing experiments, allows us to reduce the experimental burden caused by the combinatorial explosion of different datasets, languages, tasks, and models. 
    In this paper, we make two contributions to improving performance prediction for NLP tasks.
    First, we examine performance predictors not only for holistic measures of accuracy like F1 or BLEU, but also \emph{fine-grained} performance measures such as accuracy over individual classes of examples.
    Second, we propose methods to understand the \emph{reliability} of a performance prediction model from two angles: confidence intervals and calibration.
    We perform an analysis of four types of NLP tasks, and both demonstrate the feasibility of fine-grained performance prediction and the necessity to perform reliability analysis for performance prediction methods in the future.
    We make our code publicly available: \url{https://github.com/neulab/Reliable-NLPPP}    
    
\end{abstract}

\section{Introduction}
\label{sec:intro}

Performance prediction (P$^2$) aims to predict a machine learning system's performance based on features of the underlying problem, dataset, or learning algorithm.
While this topic is still relatively unexplored in the NLP context, there are a few examples of predicting performance as:
(i) a function of training or model parameters for determining the number of training iterations \cite{kolachina-etal-2012-prediction} or value of hyperparameters \cite{rosenfeld2019constructive} and identifying and terminating bad training runs \cite{domhan2015speeding}.
(ii) a function of dataset characteristics to illustrate which factors are significant predictors of system performance \cite{birch-etal-2008-predicting,turchi2008learning}, or find a subset of representative experiments to run in order to obtain plausible predictions \cite{xia-etal-2020-predicting}.
In this paper, we ask two research questions with respect to performance prediction: can we predict performance on a more \emph{fine-grained} level, and can we quantify the \emph{reliability} of performance predictions?

\begin{figure}[t]
    \centering
    \includegraphics[width=0.63\linewidth]{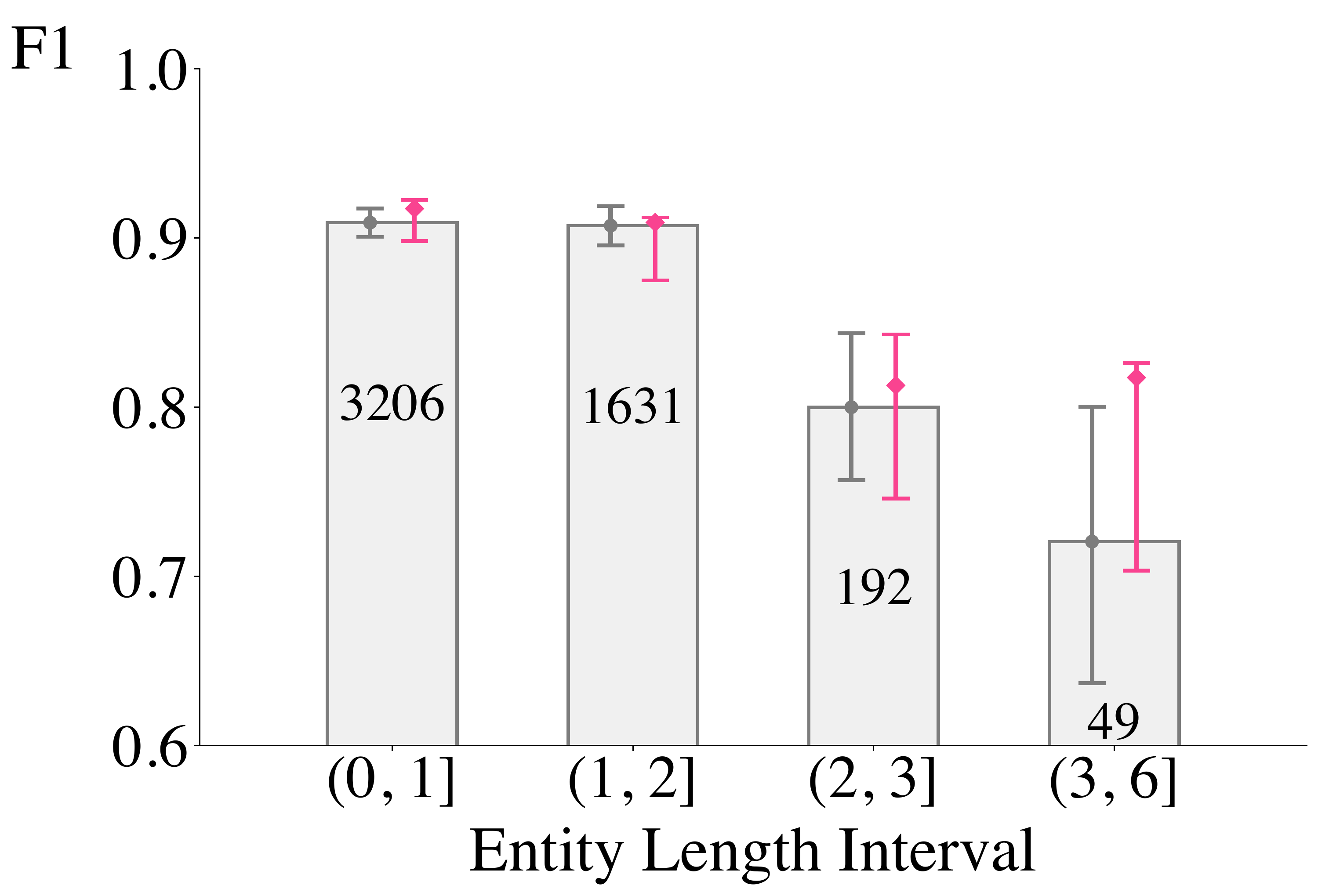}
    \vspace{-6pt}
    \caption{Breakdown of performance over different entity lengths of an NER system. \textit{Actual F1} (gray point) is calculated from actual results while \textit{predicted F1} (red point) is estimated by a performance prediction model. Gray and red lines represent corresponding confidence intervals. Numbers in each bar indicate the number of test samples in each length bucket. 
    }
    \label{fig:intro}
\end{figure}

With respect to the first contribution, previous P$^2$ methods have almost entirely focused on predicting \emph{holistic} measures of accuracy such as entity F1 \cite{ratinov2009design} or BLEU score \cite{papineni-etal-2002-bleu} over the entire dataset (\S\ref{sec:holistic}).
However, from a perspective of understanding the workings of our models, work on model analysis has demonstrated the need for more \emph{fine-grained} analysis over a wide variety of tasks \cite{kummerfeld-etal-2012-parser,kummerfeld-klein-2013-error,karpathy2015visualizing,neubig2019compare,fu-2020-interpreteval,fu2020chinese,fu2020rethinking}. 
These methods calculate separate accuracy scores for different types of examples (e.g.~accuracies for entity recognition by entity length).
Our first contribution is to examine experimental settings where we predict these fine-grained evaluation scores (\S\ref{sec:finegrained}), and also propose performance prediction methods particularly suited to this fine-grained evaluation setting (\S\ref{eq:allmodel}). 

Our second contribution is the development of methods for estimating the reliability of performance predictions.
While allowing estimation of experimental results without actually having to run the corresponding experiments may improve efficiency, if the performance predictor is wrong it may lead to missing results of a potentially important experiment.
This particularly becomes an issue when developing methods for fine-grained performance prediction, as the number of data points which can be used to predict each performance number decreases as we sub-divide datasets into finer-grained categories.
Thus, we make methodological steps towards answering two specific questions:
(i) how can we define and calculate a confidence interval over performance predictions?
(ii) how well does the confidence interval of prediction performance calibrate with the true probability of an experimental result?
Fig.~\ref{fig:intro} is an example of performance prediction and reliability analysis over fine-grained performance estimates (F1 scores over different entity length buckets) of an NER system are obtained in two ways: (i) calculated based on results from the NER system itself (in gray); (ii) estimated based on a performance prediction model, without running an actual experiment (in red).
We can observe that: (1) with fewer test samples (e.g.~\texttt{49}), confidence intervals of both actual and predicted F1 become much wider, suggesting larger uncertainty.
(2) in the last bucket, the predicted F1 (red point) is far from the actual F1 (gray point), but with a confidence interval of predicted performance (red bar), the actual F1  still falls within it, indicating the importance of knowing the level of confidence.

In experiments, we investigate the efficacy of different performance prediction models on four typical NLP tasks under both holistic and fine-grained settings, then explore methods for the reliability analysis of these performance prediction models.
Major experimental results show:
1) there is no one-size-fits-all model: best-scoring performance prediction systems in different scenarios are diverse. In particular, one of our proposed models achieved the best results on the Part-of-Speech task (\S\ref{sec:eval-pp}).
2) a better performance prediction model doesn't imply better calibration (\S\ref{sec:eval-reliability}).
3) all four performance prediction models 
(including previous top-scoring ones) produce confidence intervals over-confidently (\S\ref{sec:eval-reliability}).
\vspace{-2pt}
\section{Performance Prediction: Formulation and Applicable Scenarios}
\vspace{-3pt}
In this section, we will mathematically define performance prediction and its application in the holistic and fine-grained evaluation.
\subsection{Formulation}

Given a machine learning model $\mathcal{M}$, which is trained over a training set $\mathcal{D}^{tr}$ based on a specific training strategy $\mathcal{S}$, we then test the dataset $\mathcal{D}^{ts}$ under evaluation setting $\mathcal{E}$ and the test result $y$ can be formulated as a function of the following inputs:
\begin{align}
    y = f(\mathcal{M}, \mathcal{D}^{tr}, \mathcal{D}^{ts}, \mathcal{S}, \mathcal{E}), \label{eq:true}
\end{align} This we will refer to as the \emph{actual} performance (e.g.,~F1 score), which requires us to run an actual experiment.

Alternatively, to calculate $ y$, instead of performing a full training and evaluation cycle, one can directly estimate it by extracting features of $\mathcal{M}$, $\mathcal{D}^{tr}$, $\mathcal{D}^{ts}$, $\mathcal{S}$, and running them through a prediction function
\begin{align}
    \hat{y} = g(\Phi_{\mathcal{M}}, \Phi_{\mathcal{D}^{tr}}, \Phi_{\mathcal{D}^{ts}}, \Phi_{\mathcal{S}}, \mathcal{E};\Theta), \label{eq:full}
\end{align}
where $\Phi(\cdot)$ represents features of the input, and $\Theta$ denotes learnable parameters.
We will refer to this as our \emph{predicted} performance.
As long as Eq.~\ref{eq:full} is fast to calculate and a relatively accurate approximation of Eq.~\ref{eq:true}, it allows us to get a reasonable idea of expected experimental results much more efficiently than if we had to actually experiment.


In a real scenario, not all inputs in Eq.~\ref{eq:full} need to be taken into account, and researchers can adopt different inputs for a particular use.
For example, \citet{domhan2015speeding} define $\hat{y}$ as a function of training strategy $\mathcal{S}$ (e.g., different hyper-parameter settings) so that they can know which training setting can lead to bad performance without running.
\citet{Dodge2020} estimate validation performance as a function of computation budget  to conduct more robust model comparisons. 

\paragraph{Why Performance Prediction matters for NLP tasks}
Firstly, for some NLP tasks with few resources, it is challenging to build and test systems for all languages or domains. For example, the task of Machine Translation (MT) for low resource languages is hard due to the lack of the large parallel corpora, preventing us from measuring system performance in these scenarios \cite{xia-etal-2019-generalized,xia-etal-2020-predicting}. 
Therefore, performance prediction is useful in that it can efficiently and comprehensively give insights about the workings of models over a wide variety of task settings.
Secondly, performance prediction can be used to alleviate the data sparsity problem in fine-grained evaluation, which plays an important role in current NLP task evaluation \cite{fu-2020-interpreteval}.

In this paper, we consider two performance prediction scenarios, a holistic evaluation setting that most previous works have explored, and a novel setting of predicting fine-grained evaluation metrics. Below, we briefly describe them.


\subsection{Holistic Evaluation}  \label{sec:holistic}

%

Performance prediction in holistic evaluation aims to estimate an overall score (e.g., BLEU) based on dataset characteristics, specifically,
\vspace{5pt}
\begin{align}
    \hat{y} = g^{\text{holistic}}(\Phi_{\mathcal{D}^{tr}}, \Phi_{\mathcal{D}^{ts}};\Theta), \label{eq:low}
\end{align} 
\noindent
where $\Phi(\cdot)$ represents features of input and $\Theta$ denotes learnable parameters.

\paragraph{Featurization}
In practice, we choose a machine translation (MT) task and a Part-of-Speech task (POS) task in this setting.
We use the same set of dataset features as \cite{xia-etal-2020-predicting}, including the language features and the source and the target, or transfer language.


\subsection{Fine-grained Evaluation} \label{sec:finegrained}
In contrast, fine-grained evaluation aims to break down the overall score into different interpretable parts, allowing us to identify the strengths and weakness of learning systems. 
For example, the accuracy of an NER system with an overall F1 score $90$ ($\%$) can be partitioned into four buckets based on different entity lengths $l$ (e.g., $[l=1, 1 < l \leq 3, 3 < l \leq  5, l > 5 ]$) of test entities, thereby obtaining fine-grained  F1 scores: $[93, 91, 89, 75]$, identifying that the model struggles on longer entities ($l > 5$).

Although fine-grained evaluation is advantageous in interpreting systems' performance, it frequently suffers from the data sparsity problem---a few or no test samples may be included within certain buckets. For example, in the above case it's difficult to calculate the F1 score for entities whose lengths satisfy $l > 7$ since few entities can be found in the whole test set.

With the above dilemma in mind, we define a performance prediction problem in fine-grained evaluation where the paucity of test samples in some buckets leads to an inability to compute performance accurately.

\vspace{-15pt}
\begin{align}
    \hat{y} = g^{\text{fine}}(\Phi_{M}, \Phi_{\mathcal{D}^{tr}}, \Phi_{\mathcal{D}^{ts}};\Theta), \label{eq:fine-grained}
\end{align}
where $\Phi(\cdot)$ represents features of input and $\Theta$ denotes learnable parameters.

\paragraph{Featurization}
Performing fine-grained evaluation involves two major steps: (i) partition the test set into different \textit{buckets} based on a certain \textit{aspect} (e.g., \texttt{entity length}), (ii) and calculate performance (e.g., \texttt{F1 score}) for each bucket.
Therefore, data-wise ($\Phi_{\mathcal{D}^{ts}}$), the input of performance prediction function in Eq.~\ref{eq:fine-grained} ($\mathrm{g}^{fine}(\cdot)$) can be featurized as different types of (i) buckets (ii) aspects (iii) datasets.
Additionally, we take (iv) different types of models as input.
We present brief descriptions of the above four types of features.

\noindent 1. \textit{Models}: 
We choose $12$ models for the NER task and $8$ models for the Chinese Word Segmentation (CWS) task. 
The models are built by choosing the different character encoder (e.g., ELMo~\cite{peters2018deep} and Flair~\cite{akbik2018contextual,akbikpooled}), word embedding (e.g., GloVe~\cite{pennington2014glove} and Word2Vec~\cite{mikolov2013distributed}), sentence-level encoder (e.g., LSTM~\cite{hochreiter1997long} and CNN~\cite{kalchbrenner2014convolutional}), and decoder (e.g., MLP and CRF~\cite{lample2016neural,collobert2011natural}).


\noindent 2. \textit{Datasets}: We consider $6$ ($5$) datasets for the NER (CWS) task, detailed in appendix.


\noindent 3. \textit{Attributes}: We consider the interpretable evaluation aspects proposed in works~\cite{fu-2020-interpreteval}.
We consider $9$ attributes for the NER task and $8$ attributes for the CWS task in this paper (e.g, entity length and sentence length). 


\noindent 4. \textit{Buckets}: The test entities (words) of the NER (CWS) task are partitioned into four buckets according to their attribute value. We compute the F1 score for the entities.

\section{Parameterized Regression Functions} \label{eq:allmodel}
The performance prediction model takes in a set of features that characterize an experiment's peculiarities and predict performances based on different parameterized regressors $\mathrm{g}(\cdot)$ in Eq.~\ref{eq:full}. We first describe methods explored by previous works and then present a tensor regression-based approach that is particularly well-suited for fine-grained performance prediction.


\subsection{Gradient Boosting Methods}
Previous work on performance prediction has used gradient boosted decision tree models \cite{ganjisaffar2011bagging,DBLP:journals/corr/ChenG16},
which demonstrate robust performance on the relatively low-data scenarios we often encounter in performance prediction tasks. We specifically explore the following two models:

\noindent \textbf{XGBoost} \cite{DBLP:journals/corr/ChenG16} is a tree boosting system widely used to solve problems such as ranking, classification, and regression. We use the same experimental setting as described in \cite{xia-etal-2020-predicting}.

\noindent \textbf{LightGBM} \cite{Ke2017} is a gradient boosting framework. Compared with XGBoost, which utilizes a level-wise tree growth in the decision tree, LightGBM uses a leaf-wise splitting method.

\subsection{Tensor Regression} 
Besides gradient boosted trees, we also present tensor regression-based performance prediction models.
Tensors are multidimensional arrays that can concisely depict the structure of the data. The order of a tensor is its number of dimensions.
For example, in the NER task, the four feature dimensions of a tensor are model, dataset, attribute and bucket, with each slice representing one underlying relationship between the two dimensions.
Applying tensor factorization algorithms in the performance prediction setting allows us to determine the interdependencies between multiple aspects of the tasks simultaneously.

\begin{figure}[t!]
    \centering
    \includegraphics[width=0.8\linewidth]{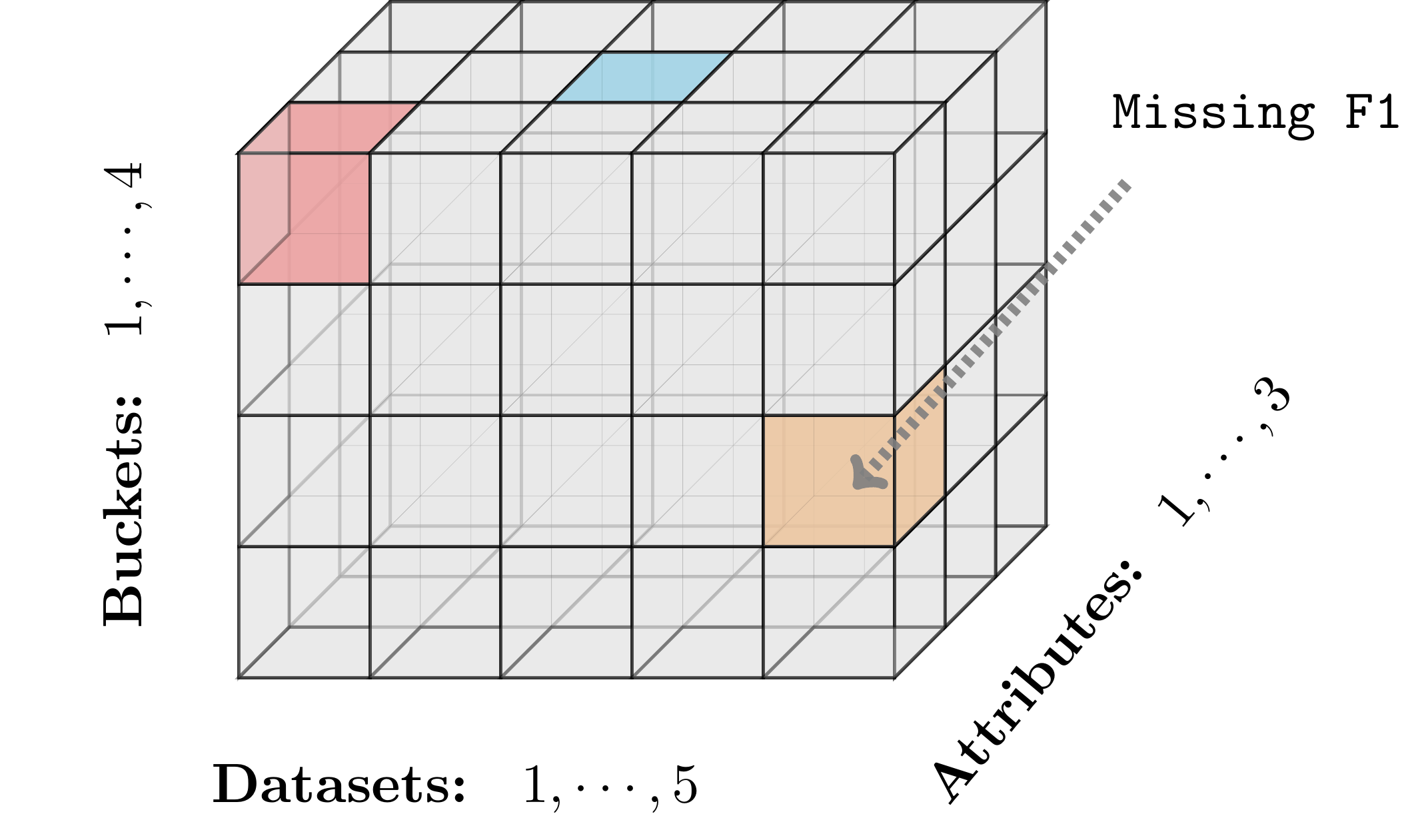}
    \vspace{-7pt}
    \caption{Illustration of performance tensor in the fine-grained evaluation scenario. Colored entries represent missing performances that would be predicted.
    }
    \label{fig:tensorF1-2}
\end{figure}

\paragraph{Performance Prediction as Tensor Completion}
To formulate the performance prediction task as a tensor regression problem: (i) we first define a \textit{performance tensor} that each entry stores a performance value under a specific setting determined by input features (described in \S\ref{sec:finegrained}); (ii) missing entries in performance tensor can be predicted based on different tensor completion techniques.

Specifically, taking fine-grained evaluation for example, 
we define a fine-grained performance tensor as $\mathbf{Y} \in \mathbb{R}^{I_1 \times I_2 \times I_3 \times I_4}$, 
where $\mathbf{Y}_{ijkt}$ denotes the performance (e.g.~F1 score) of the $i$-th model (e.g.~\texttt{BERT-based Tagger}) on the $j$-th bucket (e.g.~\texttt{$2_{nd}$}) that is obtained by partitioning the $k$-th dataset (e.g.~\texttt{CoNLL03}) based on the $t$-th attribute (e.g.~\texttt{entity length}). $I_1, I_2, I_3, I_4$ denote the number of models, buckets, datasets, attributes. 
Fig.~\ref{fig:tensorF1-2} elaborates on this, in which three dimensions (buckets, datasets, and attributes) are considered for the sake of presentation.

\noindent \textbf{CP Decomposition}
The CP decomposition \cite{doi:10.1002/sapm192761164} expresses a tensor $\mathbf{Y}$ as a sum of lower rank tensors. For example, an order 4 tensor can be decomposed as the sum of $R$ rank-1 tensors, each being the outer product of four vectors in each dimension.


\noindent \textbf{Robust PCA} Robust PCA is a modification of principal component analysis (PCA) \cite{DBLP:journals/corr/abs-0912-3599}.
If a tensor can be conceived as a superposition of low-rank components and a sparse component, Robust PCA attempts to recover the low-rank and sparse components. The sparse components can be considered as the gross, but sparse noise in the dataset. 




\section{Statistical Preliminaries}
Before going into our second contribution to establishing reliability of performance prediction, we describe two relevant concepts from statistics.



\subsection{Confidence Interval (CI)} \label{se:ci}
The confidence interval (CI) is a range of possible values for an unknown parameter associated with a \textit{confidence level} of $\gamma$ 
\cite{nakagawa2007effect,dror-etal-2018-hitchhikers} that the actual parameter can fall into the suggested range. Specifically, suppose that we are interested in estimating the underlying true parameter of $\omega$. Given an observed parameter estimate of $\hat{\omega}$, obtained from the data, we aim to compute an interval with a confidence level $\gamma$ that $\omega$ lies in an interval CI.

Commonly, there are two approaches to calculate confidence intervals, depending on our knowledge about the distribution of the statistics of interest. When an analytical form exists and we have reasonable assumptions on the distribution, we can employ the normal theory or use Student's t-distribution to construct a confidence interval.

Regarding data drawn from a completely unknown distribution, a CI can be calculated by a bootstrapping method \cite{efron1992bootstrap,johnson2001introduction}. The main idea behind the bootstrapping method is to simulate the real distribution by sampling with replacement from a distribution that approximates it, thereby allowing us to make inferences about the statistics of interest and construct confidence intervals. Common methods to construct the CI with bootstrap include the percentile method, where after specifying a confidence level $\gamma$, we take the range of points that cover the middle $\gamma$ proportion of bootstrap sampling distribution $\hat{Y}$ as the desired confidence interval, represented by $(Q_{\hat{Y}((1-\gamma)/2)}, Q_{\hat{Y}((1+\gamma)/2)})$, where $Q$ denotes the quantile. Works on establishing confidence for results in NLP tasks using this bootstrap method include \citet{Koehn2004} and \citet{li2017modeling}.

\subsection{Model Calibration (MC)} \label{sec:calibration}


Calibration \cite{gleser1996measurement}, also known as reliability, refers to the ability of a model to make good probabilistic predictions. For a discrete distribution over events, a model is said to be well-calibrated if for those events that the model assigns a probability of $p$, the long-run proportion that the event actually occurs turns out to be $p$. For example, if a weather forecast model predicts that there is a 0.1 probability of rain at 7 a.m., then when observed on a large number of random trials at 7 a.m., the model is well-calibrated if 0.1 of them actually do result in rain. Similarly, for a classification model matching the probability a model assigns to a predicted label (i.e., \textit{confidence}) and the correctness measure of the prediction (i.e., \textit{accuracy}) \cite{wang-etal-2020-inference} is desired.

Nonetheless, it is common that a model could have a high predictive accuracy, but poor calibration if the model systematically over- or under-estimates its confidence in the predictions it makes.
One way to quantify miscalibration is to use \textit{Expected Calibration Error} (ECE; \citet{Naeini2015}), which aims to quantitatively characterize the difference in expectation between confidence and accuracy.
To calculate ECE, the predictions should first be partitioned into $M$ buckets based on the confidence of the predictions, where $N$ represents the total number of prediction samples and $|B_{m}|$ is the number of samples in the $m$-th bucket.
Given these buckets, ECE can be defined as, 

\vspace{-15pt}
\begin{align}
    \mathrm{ECE} = \sum_{m=1}^{M}\frac{|B_m|}{N}|\mathrm{acc}(B_m) - \mathrm{conf}(B_m)|,
\end{align}

\vspace{-5pt}
\noindent
where $\mathrm{acc}(B_m)$ denotes the accuracy of $B_m$,

\vspace{-15pt}
\begin{align}
    \mathrm{acc}(B_m) = \frac{1}{|B_m|} \sum_{i\in B_m}\mathbf{1}(\hat{y}_i = y_i),
    \label{eq:acc}
\end{align}

\vspace{-8pt}
\noindent
where $\hat{y}$ and ${y}$ represent predicted and ground truth labels respectively.
$\mathrm{conf}(B_m)$ represents the average confidence of bucket $B_m$,

\vspace{-15pt}
\begin{align}
    \mathrm{conf}(B_m) = \frac{1}{|B_m|}\sum_{i\in B_m}\hat{p}_i.
    \label{eq:conf}
\end{align}

\vspace{-8pt}
\noindent
where $\hat{p}$ represents the prediction confidence of sample $i$.







\section{On Reliability of $P^2$ Models}

Now we discuss our methodology for predicting the reliability of performance prediction models through confidence intervals and the calibration of those confidence intervals. 

\subsection{CIs of Predicted Performance} \label{sec:ci}

We refer to $y \sim Y$ as an actual observed performance as in Eq.~\ref{eq:true} for a specific task (e.g., NER). $y$ is the output of an NLP system learned on a dataset $\mathcal{D} = (\mathcal{D}^{tr}, \mathcal{D}^{ts})$.
We refer to $\hat{y} \sim \hat{Y}$ as a predicted performance estimated as Eq.~\ref{eq:full}. $\hat{y}$ is the output of a performance prediction model learned from a dataset $\Phi(\mathcal{D}) = (\Phi(\mathcal{D}^{tr}), \Phi(\mathcal{D}^{ts})) $, where $\Phi(\cdot)$ represents the input dataset features.
Our goal is to compute a confidence interval w.r.t a predicted performance $\hat{y}$, to make inference about $Y$. 
\paragraph{Bootstrap for CI of Predicted Performance}

One potential challenge is that we cannot make plausible assumptions about the distribution of predicted performances $\hat{Y}$, which prevents us from using popular parametric methods (as mentioned in \S~\ref{se:ci}) to calculate the confidence interval.
Instead, we resort to a bootstrap resampling method as adopted in \cite{efron1992bootstrap}, to simulate $\hat{Y}$.


To achieve this, we first
(i) sample different training sets for the performance prediction model $\Phi(\mathcal{D})^{tr}_1, \Phi(\mathcal{D})^{tr}_2, \cdots, \Phi(\mathcal{D})^{tr}_K \sim \Phi(\mathcal{D})^{tr}$, and then 
(ii) train $K$ performance prediction models using Eq.~\ref{eq:full} on each of the $K$ partitions,
and (iii) evaluate $K$ models on $\Phi(\mathcal{D})^{ts}$, thereby obtaining a prediction distribution $\hat{Y}$.
 From this resampling distribution, we use the percentile method, taking the top $(1-\gamma)/2$ and the bottom $(1+\gamma)/2$ of the distribution as higher and lower bounds for the confidence interval.

\subsection{Calibration of CI}

Because we calculate confidence intervals of the predicted performance $\hat{y}$, drawn from the distribution of $\hat{Y}$, rather than the actual ${y}$ from $Y$, it's still unclear if our predicted CI is reliable enough to cover the actual performance. 
In other words, ``from an infinite number of independent trials, does the true value \textit{actually} lie within the intervals approximately 95\% of the times?"

To answer this question, we establish a method to measure calibration for the confidence interval of predicted performance. 
To check (i) if $y$ could be generally contained in the prediction intervals reasonably well, and (ii) if a prediction model produces predictions that are not over or under-confident, we empirically examine the prediction distributions and establish the reliability of the confidence intervals.


To this end, we extend the definition of calibration in classification setting to our regression problem.
Specifically, we formulate {confidence level} $\gamma$ as prediction confidence $\mathrm{conf}$ defined in Eq.~\ref{eq:conf}, and then the original definition of different $M$ buckets can be instantiated as different confidence levels here: $\gamma_1, \cdots, \gamma_M$. 
The \emph{accuracy} at each confidence level $\gamma_b$ 
defined as follows:

\vspace{-15pt}
\begin{align}
    \mathrm{acc}(\gamma_b) = \frac{\sum_{i=1}^N \mathbbm{1} (A < {y_i} <  B) }{N},  \label{eq:correctness}
\end{align}

\vspace{-5pt}
\noindent
where $i\in [1,N]$, $b \in [1, M]$. $N$ represents the number of test samples.
$y_i$ denotes the actual performance for the test sample $i$.
$A = (Q_{\hat{Y}((1-\gamma_{b})/2)}$ and  $B = Q_{\hat{Y}((1+\gamma_{b})/2)})$.

\begin{figure}[t!]
    \centering
    \includegraphics[width=0.95\linewidth]{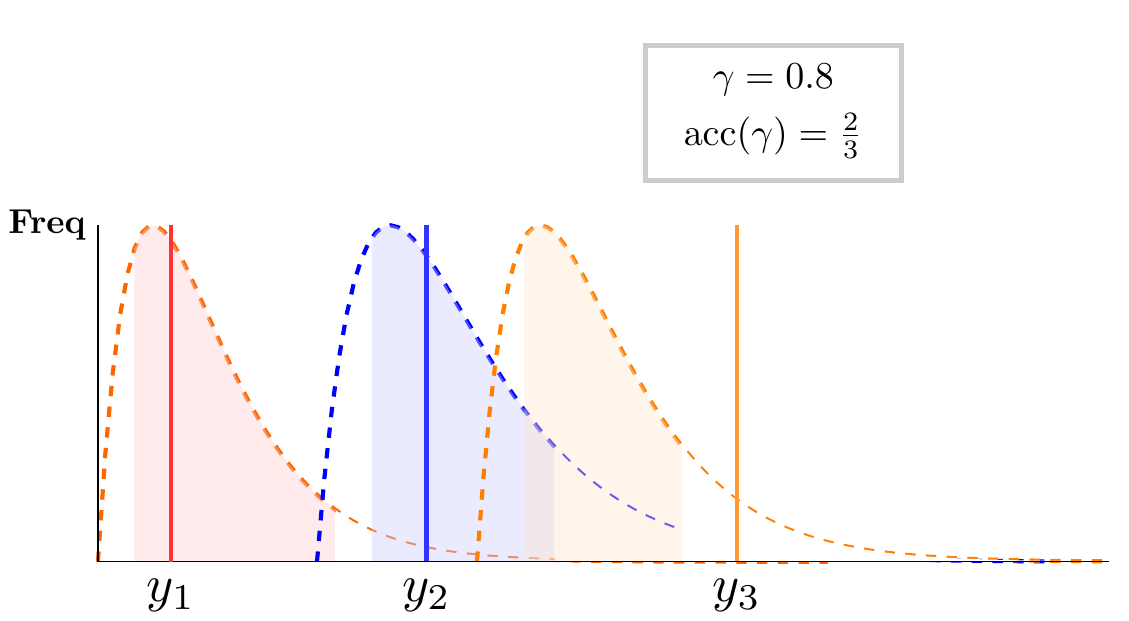}
    \vspace{-7pt}
    \caption{Illustration of calibration on confidence intervals ($\gamma_{b} = 0.8$) w.r.t.~the predicted performance. Solid lines represent actual performances ($y$) while dashed lines denote resampled predicted performances using bootstrap. Boundaries of shaded areas indicate confidence interval $(Q_{\hat{Y}((1-\gamma_{b})/2)}, Q_{\hat{Y}((1+\gamma_{b})/2)})$. Intuitively, $y_1$ and $y_2$ fall into confidence intervals of corresponding bootstrapped distributions $\hat{Y_1}$ and $\hat{Y}_2$.
    }
    \label{fig:tensorF1}
\end{figure}

Intuitively, $\mathrm{acc}(\gamma_{b})$ represents the relative frequency of the actual value $y$ falling into the predicted confidence interval w.r.t.~$\hat{y}$.
Fig.~\ref{fig:tensorF1} illustrates how $\mathrm{acc}(\gamma_{b})$ is calculated:
given three samples whose performances are to be predicted, the denominator of $\mathrm{acc}(\gamma_{b}=0.8)$ is $3$ while the numerator tallies how many times ($2$ in this case) the actual performances (i.e.,~$y_1, y_2, y_3$) of three samples fall into the confidence interval (with $\gamma_{b}=0.8$) of corresponding bootstrapped distributions.

Based on Eq.~\ref{eq:correctness}, we can re-write a calibration error $\mathrm{CE}$ as: 
\vspace{-20pt}

 \begin{align}
  \mathrm{CE} &= \sum_{b=1}^M \mid \mathrm{acc}(\gamma_{b}) - \gamma_{b} \mid \\
   &= \sum_{b=1}^M \mid \frac{\sum_{i=1}^N \mathbbm{1} (A < {y_i} <  B) }{N} - \gamma_{b} \mid
   \label{eq:CE}
\end{align}
\vspace{-15pt}
 
\section{Experiments}
In this section, we break down our experimental results into answering two research questions sections: (1) how well do our underlying performance predictors work, particularly the newly proposed tensor-based predictors and on the newly proposed task of fine-grained performance prediction? (2) how well can we estimate the reliability of our performance predictions?

\paragraph{Models}

Besides the four performance prediction models (CP, PCA, XGBoost, LGBM) that we have introduced in \S\ref{eq:allmodel}, following \citet{xia-etal-2020-predicting}, we additionally use a simple mean value baseline model which predicts an average of scores $s$ from the training folds for all test entries in the left-out evaluation fold:

\vspace{-10pt}
\begin{align}
  \hat{s}_{\text{mean}}^{(i)} = \frac{1}{|\mathcal{D}\setminus \mathcal{D}^{(i)}|}\sum_{s\in \mathcal{D}\setminus \mathcal{D}^{(i)}}s;i\in1,...k,
\end{align}

\vspace{-8pt}
\noindent
where $\mathcal{D}^{(i)}$ is the left-out data used to evaluate the model performance. 



\paragraph{Hyper-parameters}

Detailed information about the hyper-parameters used in training the performance prediction models in various tasks is provided in the appendix.


\paragraph{Tasks}
We explore performance prediction on four tasks:
(1) Machine Translation (MT) \cite{DBLP:journals/corr/abs-1907-05791},
(2) Part-of-Speech tagging (POS),
(3) Named Entity Recognition (NER),
(4) Chinese Word Segmentation (CWS). 
To compare the performance of tensor-based models and gradient boosting models on the same dataset, we convert the datasets used in different prediction tasks to tensors. 
Statistics of the tensor data are shown in the appendix.

\subsection{Evaluation of Performance Prediction} \label{sec:eval-pp}
\paragraph{Setup}
To investigate the effectiveness of the performance prediction models across different tasks, we conduct $k$-fold cross-validation for evaluation. Specifically, we randomly partition the entire experimental data $\mathcal{D}$ into $k=5$ folds, use $4$ folds for training, and test the model's performance on the remaining fold. To evaluate the result, we calculate the average root mean square error (RMSE) between the predicted scores and the true scores. 

\paragraph{Results}
The RMSE scores of different performance tasks are shown in Tab.~\ref{tab:result-RMSE}. Notably, RMSE scores across different tasks should not be compared directly, because the scales of the evaluation metrics are different. 
We observed that:

(1) Overall, all four models we investigated outperform the baseline by a large margin, indicating their effectiveness on these four performance prediction tasks.
(2) Comparing two tensor-based models, PCA consistently outperforms CP. Notably, our proposed tensor regression model (PCA) has surpassed the previous best-performing system (XGBoost \cite{xia-etal-2020-predicting}) on the \texttt{POS} dataset and achieved comparable result on the \texttt{MT} dataset despite the relatively high sparsity of the tensor (0.346).
(3) We observe that CP achieves much worse performance on the \texttt{POS} dataset. One potential reason is that: CP is sensitive to datasets (like \texttt{POS}) that exhibit large variance along some feature dimensions, which can not be alleviated by feature scaling.
(4) There is no one-size-fits-all model: on different datasets, the corresponding best-scoring performance prediction models are diverse, suggesting that we should take dataset's characteristics into account when selecting a model for a specific performance prediction scenario.

\begin{figure}[htb!]
    \centering
    \includegraphics[width=0.77\linewidth]{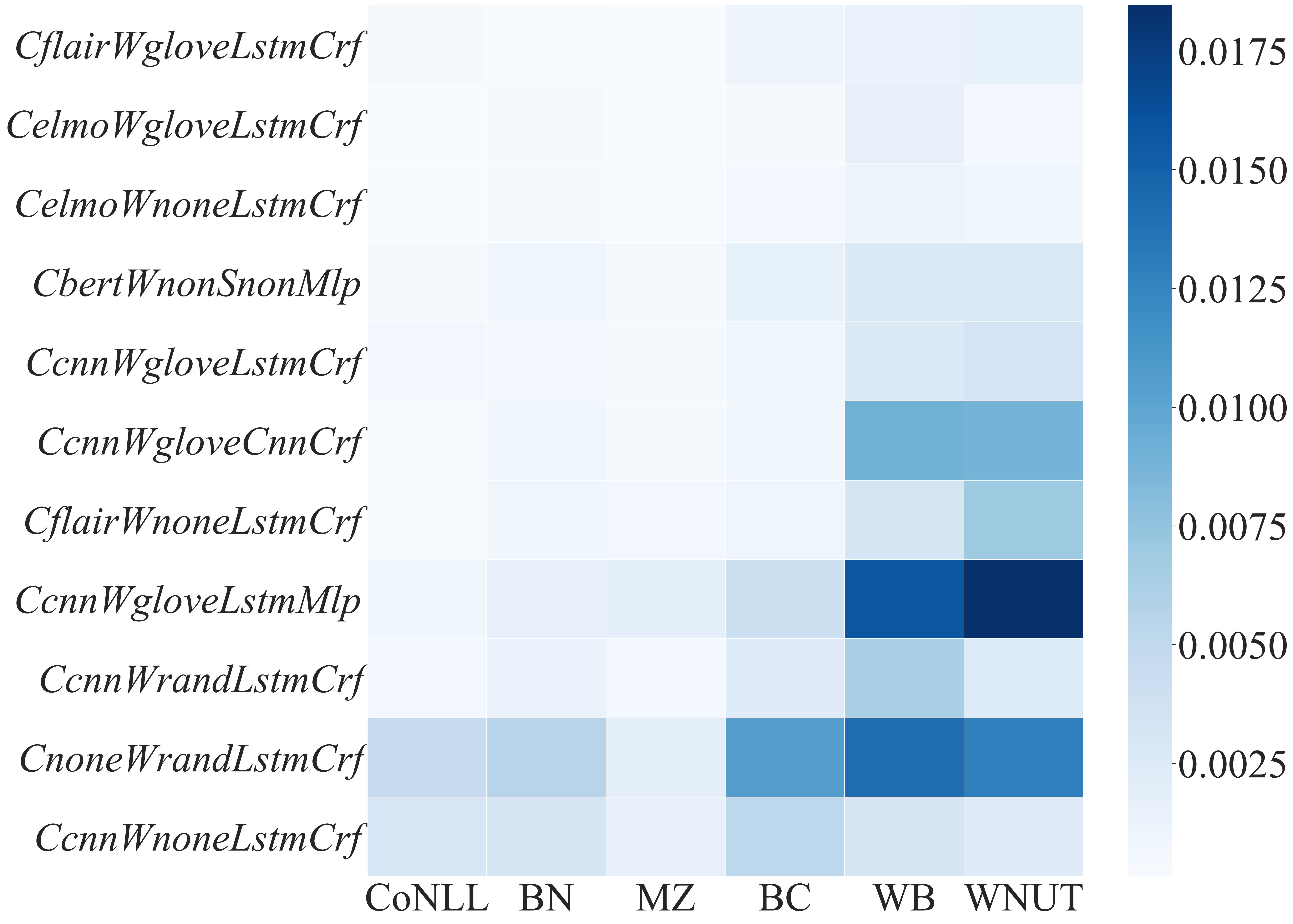}
    \vspace{-6pt}
    \caption{Each grid in the heatmap denotes the mean square residual fixed on the corresponding model (y-axis) and dataset (x-axis) aggregated over all attributes and buckets. The colorbar on the right denotes the value of the mean square residual. Readers can refer to this work~\cite{fu-2020-interpreteval} to get more details about the information of models and attributes.
    }
    \label{fig:perf_analysis_md}
\end{figure}

\begin{figure}[htb!]
    \centering
    \includegraphics[width=0.7\linewidth]{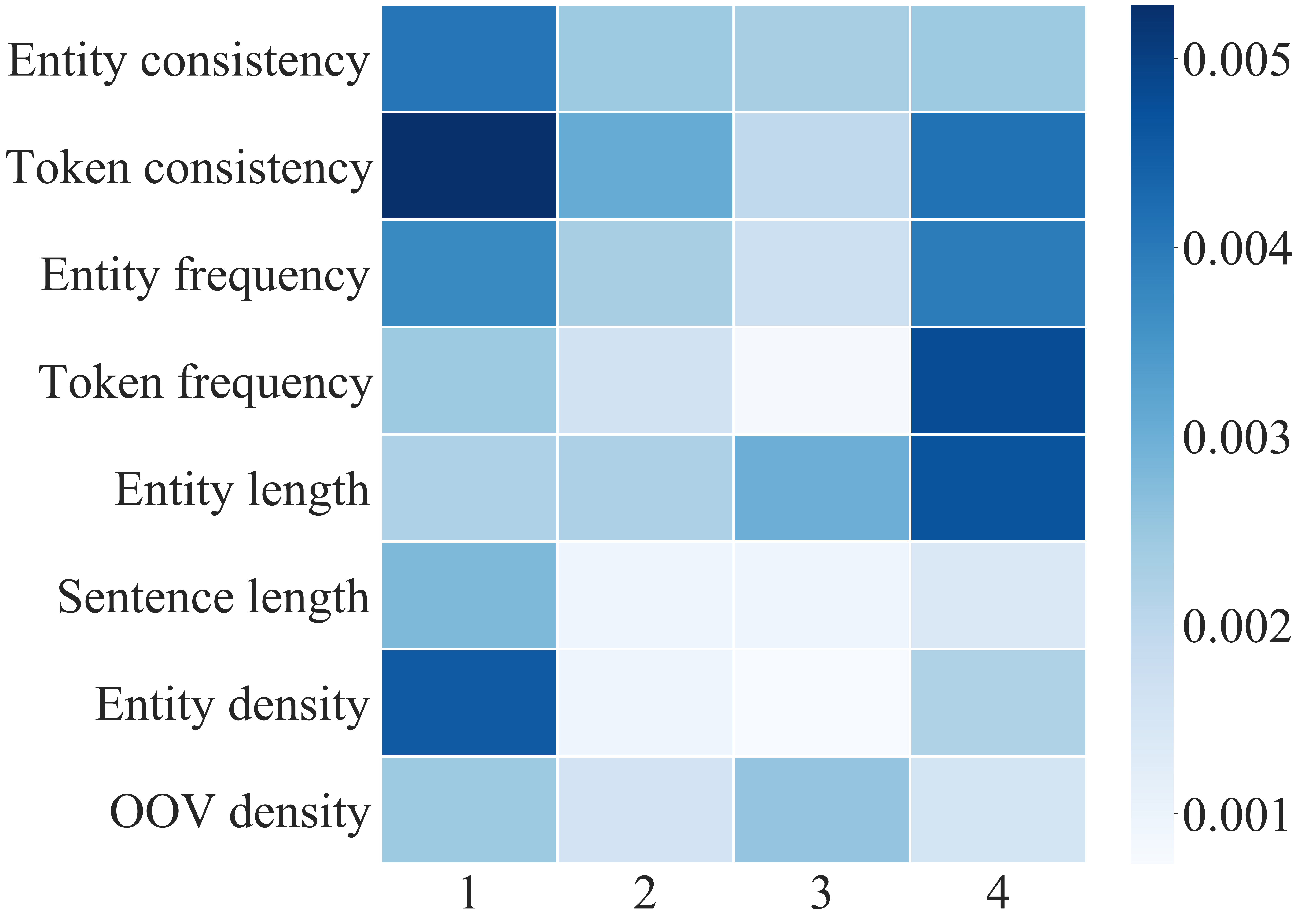}
    \vspace{-6pt}
    \caption{Each grid denotes the mean square residual fixed on the corresponding attribute (y-axis) and bucket (x-axis) aggregated over all models and datasets. Buckets 1 to 4 are ordered in increasing attribute value.
    }
    \label{fig:perf_analysis_ab}
\end{figure}

\paragraph{Prediction Error Analysis}
In \S\ref{sec:intro} and Fig.~\ref{fig:intro}, we reveal how entities with different lengths influence the performance prediction, a result of the underlying paucity of data. Here we perform a more detailed error analysis to understand the factors that influence the performance of performance prediction models. Specifically, we perform a case study on the \texttt{NER} task using XGBoost and look for feature combinations on which performance predictions show poor results. 
We use XGBoost to predict ~F1 scores on all possible combinations of four feature dimensions (models, datasets, attributes, and buckets) to obtain $\hat{y}_{ijkt}$  using the combined test sets from 5-fold cross-validation. For each prediction, we calculate a square residual $(\hat{y}-y)^2$. Then, we group the square residuals by 2 of the 4 dimensions\footnote{Readers can refer to this work~\cite{fu-2020-interpreteval} to get more details.} and take their mean value aggregated over the other 2 dimensions. Fig.~\ref{fig:perf_analysis_md} shows the aggregated mean square residual (MSR) fixed on the model and dataset dimensions, and Fig.~\ref{fig:perf_analysis_ab} shows the result fixed on the attribute and bucket dimensions. In both figures, a high MSR (dark grid) means a poor performance prediction.
In Fig.~\ref{fig:perf_analysis_md}, we notice that (1) dataset-wise: \texttt{WB} and \texttt{WNUT}, and (2) model-wise: \textit{CcnnWgloveLstmMlp} and \textit{CnoneWrandLstmCrf} show poor results. We observe that (1) \texttt{WB} is generated from weblogs and \texttt{WNUT} is generated from Twitter, both of which are noisy. (2) \textit{CcnnWgloveLstmMlp} does not use a CRF-decoder, and \textit{CnoneWrandLstmCrf} does not encode character-level features, both of which are important characteristics in building an \texttt{NER} model. It is plausible that the systems have an unstable performance in those experimental settings and thus make them harder to predict. In Fig.~\ref{fig:perf_analysis_ab}, we notice that (1) a lower bucket value along the attributes \textit{entity consistency}, \textit{token consistency}, and \textit{entity density}, (2) a higher bucket value along the attributes \textit{token frequency} or \textit{entity length} lead to poor performance prediction results. In other words, the performance prediction model finds it hard to predict when there is a low label consistency of token or entity, a low entity density, and when token frequency is high and entity is long.

\renewcommand\tabcolsep{6pt}
\begin{table}[!t]
  \centering \small
    \begin{tabular}{lcccc}
    \toprule
     \multirow{2}[4]{*}{\textbf{Model}} & \multicolumn{2}{c}{\textbf{Fine-grained}} & \multicolumn{2}{c}{\textbf{Holistic}} \\
     \cmidrule(r){2-3}\cmidrule(r){4-5}
     & \textbf{NER} & \textbf{CWS} & \textbf{MT} & \textbf{POS} \\
    \midrule
    Baseline & 0.209  & 0.137  & 6.388  & 29.09  \\
    XGBoost & \textbf{0.055}  & \textbf{0.021}  & 2.463  & 7.319  \\
    LGBM  & 0.059  & 0.041  & \textbf{2.389}  & 7.673  \\
    CP    & 0.068  & 0.043  & 4.065  & 24.70  \\
    PCA   & 0.057  & 0.029  & 2.920  & \textbf{5.860}  \\
    
    \bottomrule
    \end{tabular}%
    \vspace{-7pt}
    \caption{Results (RMSE, lower scores indicate better performances) of different performance prediction models on four tasks. The lowest value of each column is bold. 
    }
      \label{tab:result-RMSE}
\end{table}%

\subsection{Evaluation of Reliability} \label{sec:eval-reliability}
\renewcommand\tabcolsep{1pt}
\begin{figure*}[htb]
  \centering \footnotesize
    \begin{tabular}{ccccc ccccc ccccc ccccc ccccc ccccccccc ccccc ccccccccccccccccccccccccccc}
   \multicolumn{5}{l}{\includegraphics[scale=0.093]{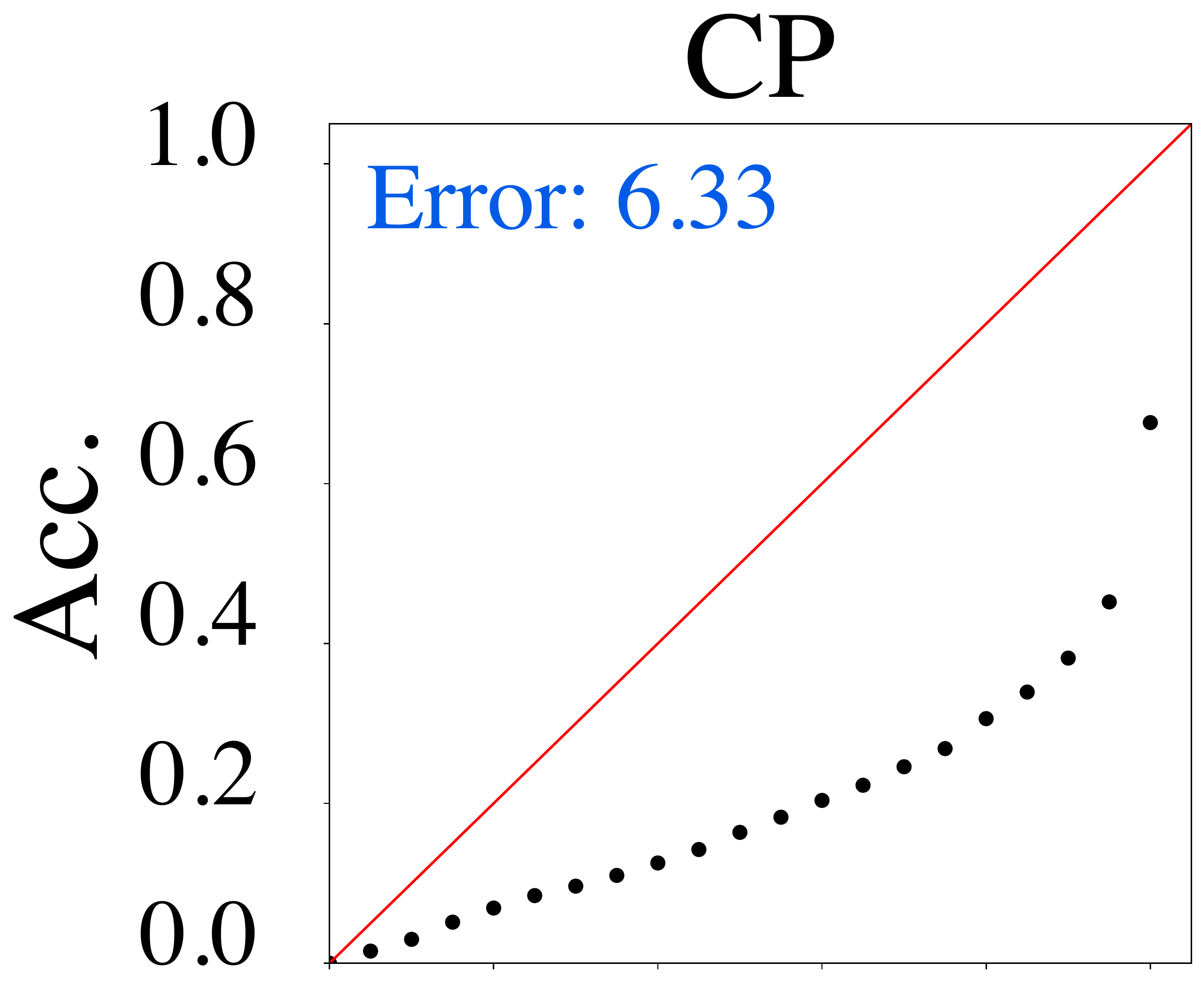}} & &&
    \multicolumn{5}{l}{\includegraphics[scale=0.092]{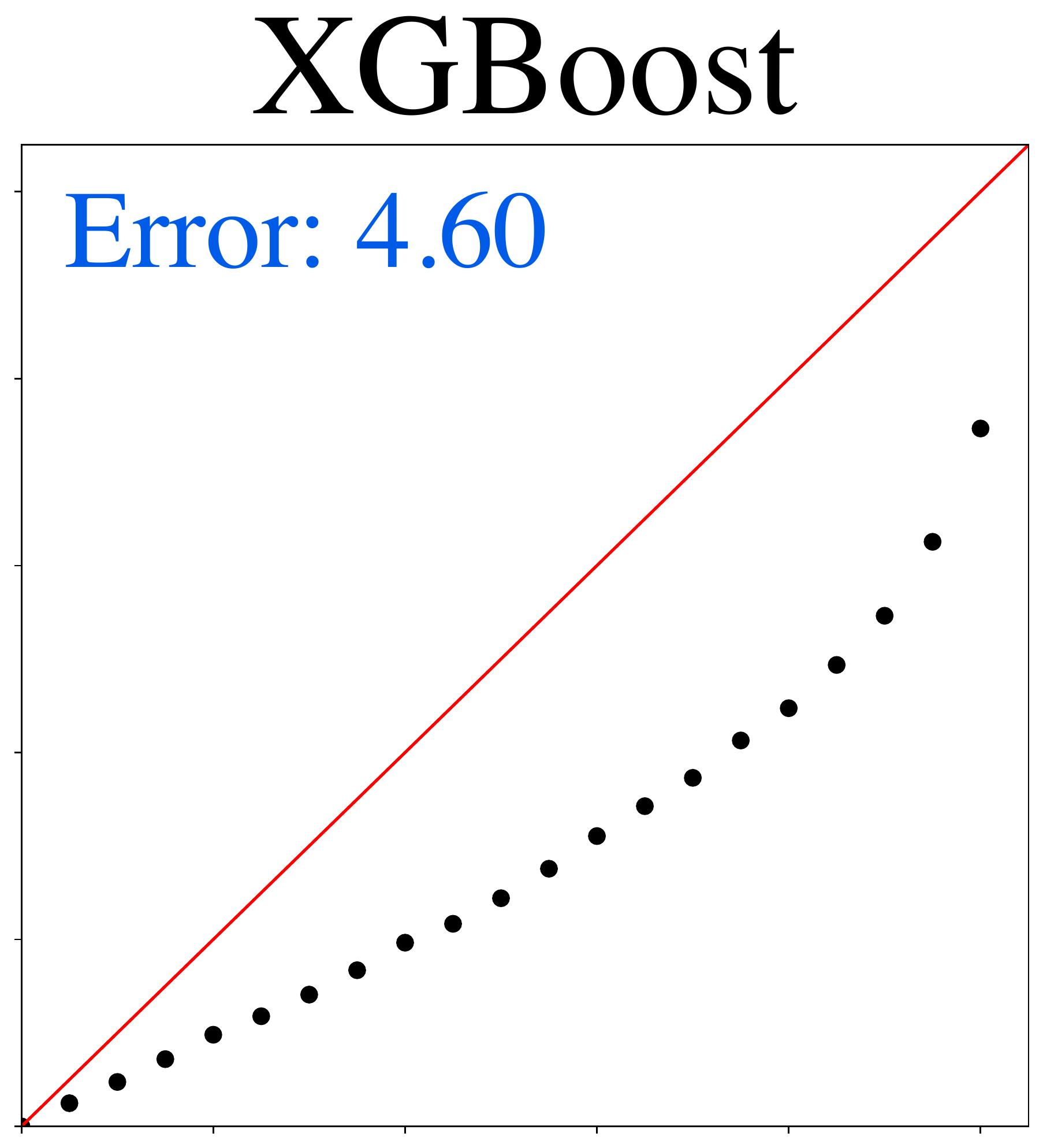}} & &&&&&&&& && \multicolumn{5}{l}{\includegraphics[scale=0.092]{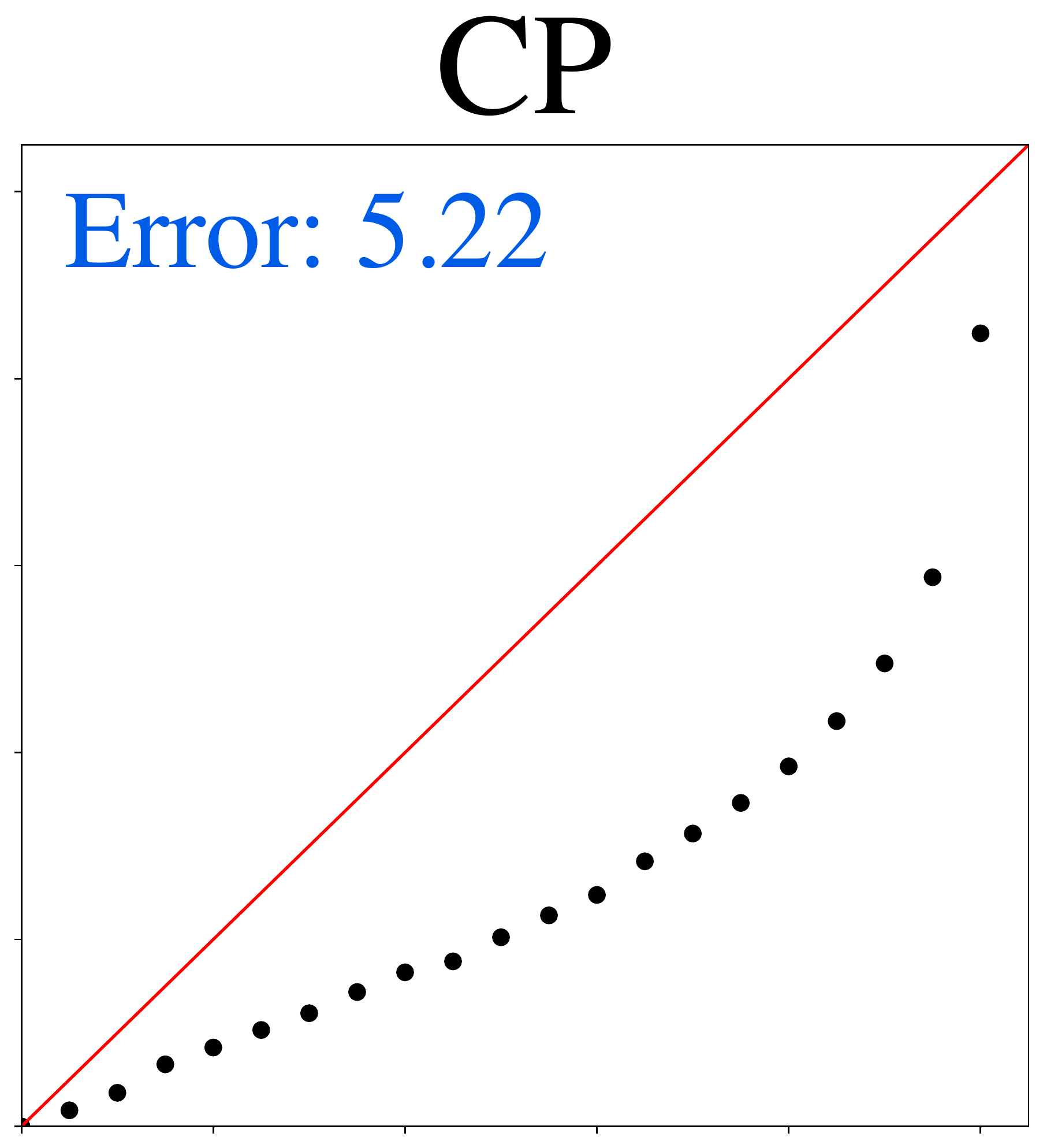}} &&&  \multicolumn{5}{l}{\includegraphics[scale=0.092]{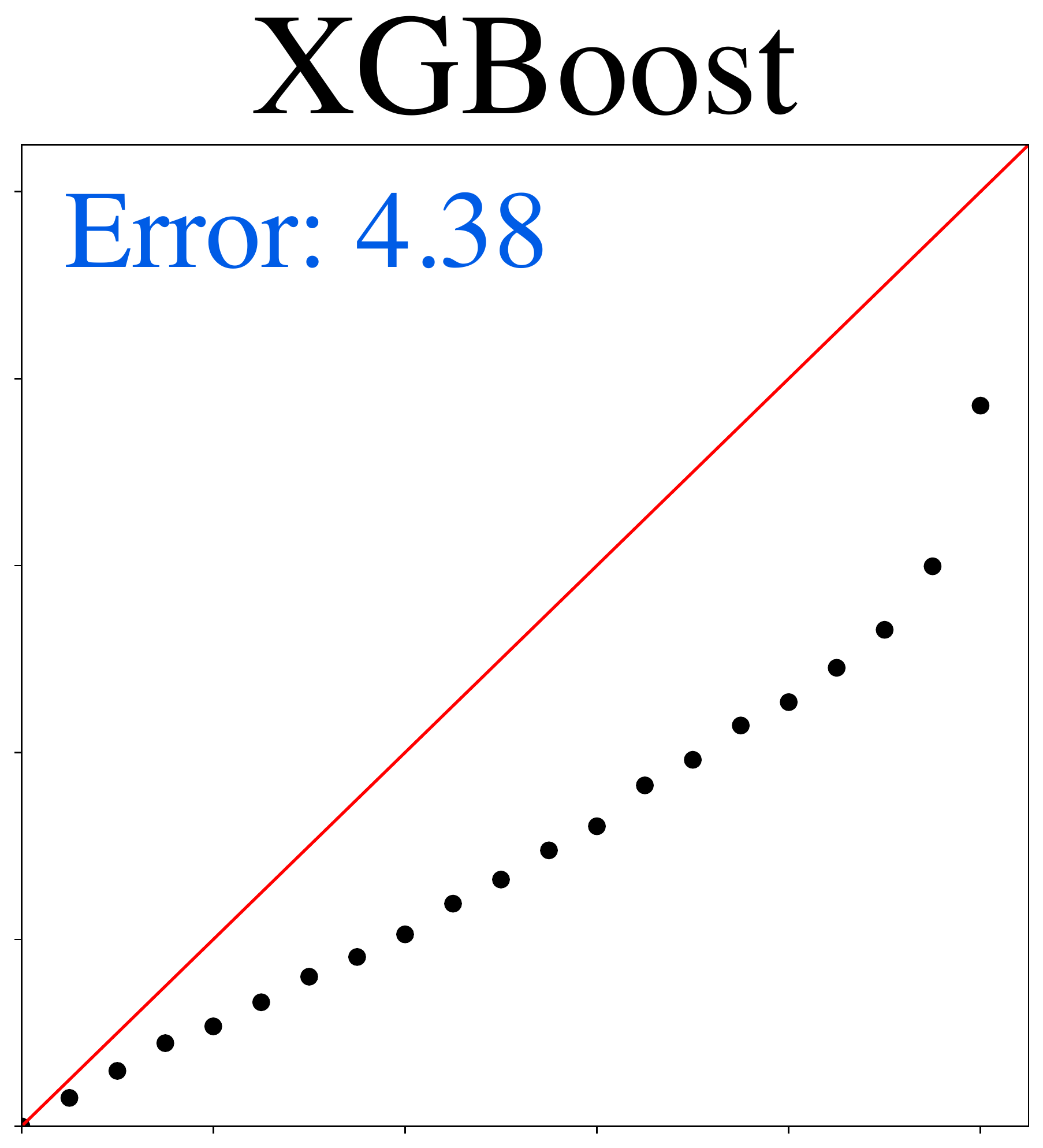}} &&&&&&&&& && \multicolumn{5}{l}{\includegraphics[scale=0.092]{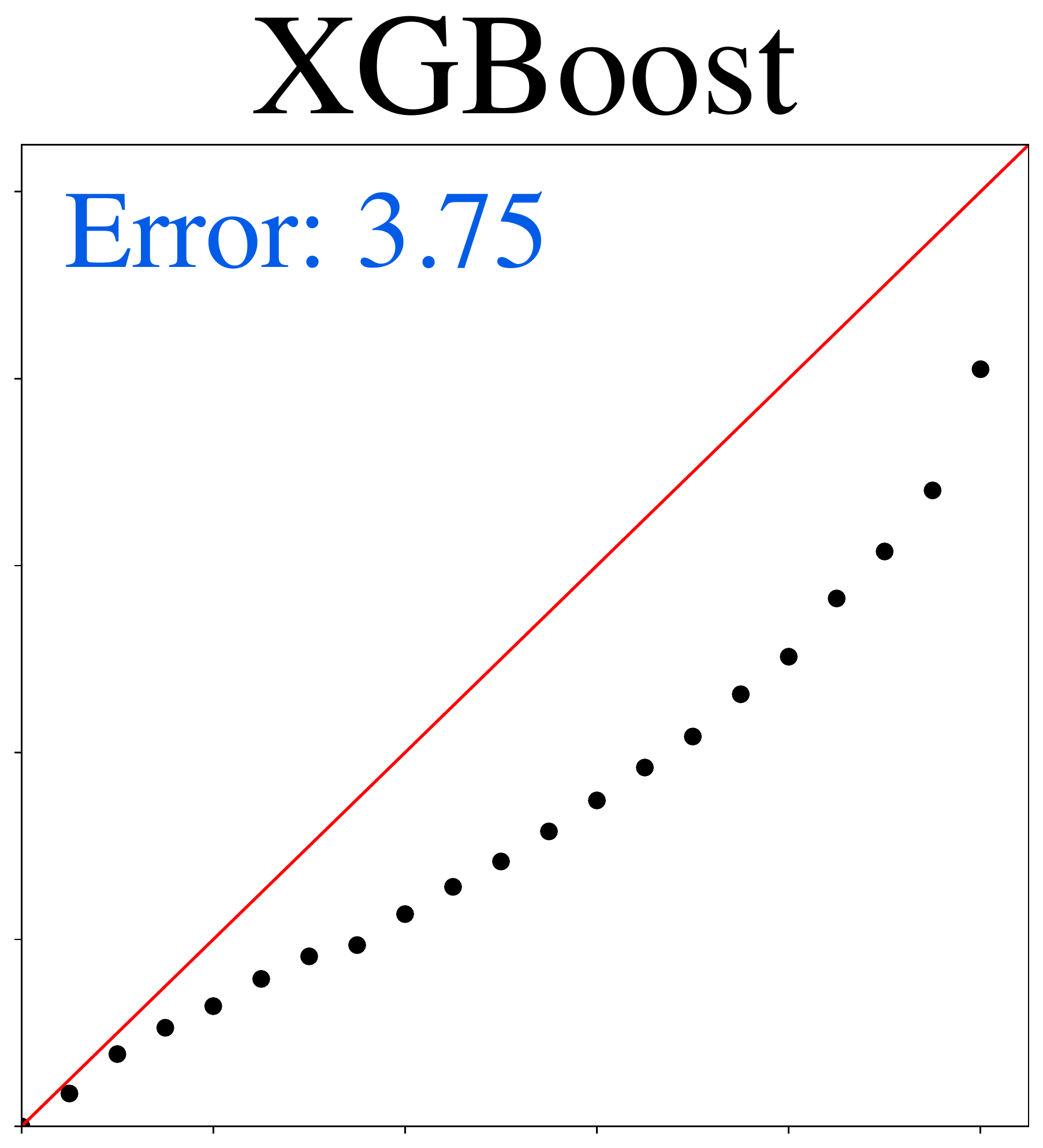}} &&&&&&&&& && \multicolumn{5}{l}{\includegraphics[scale=0.092]{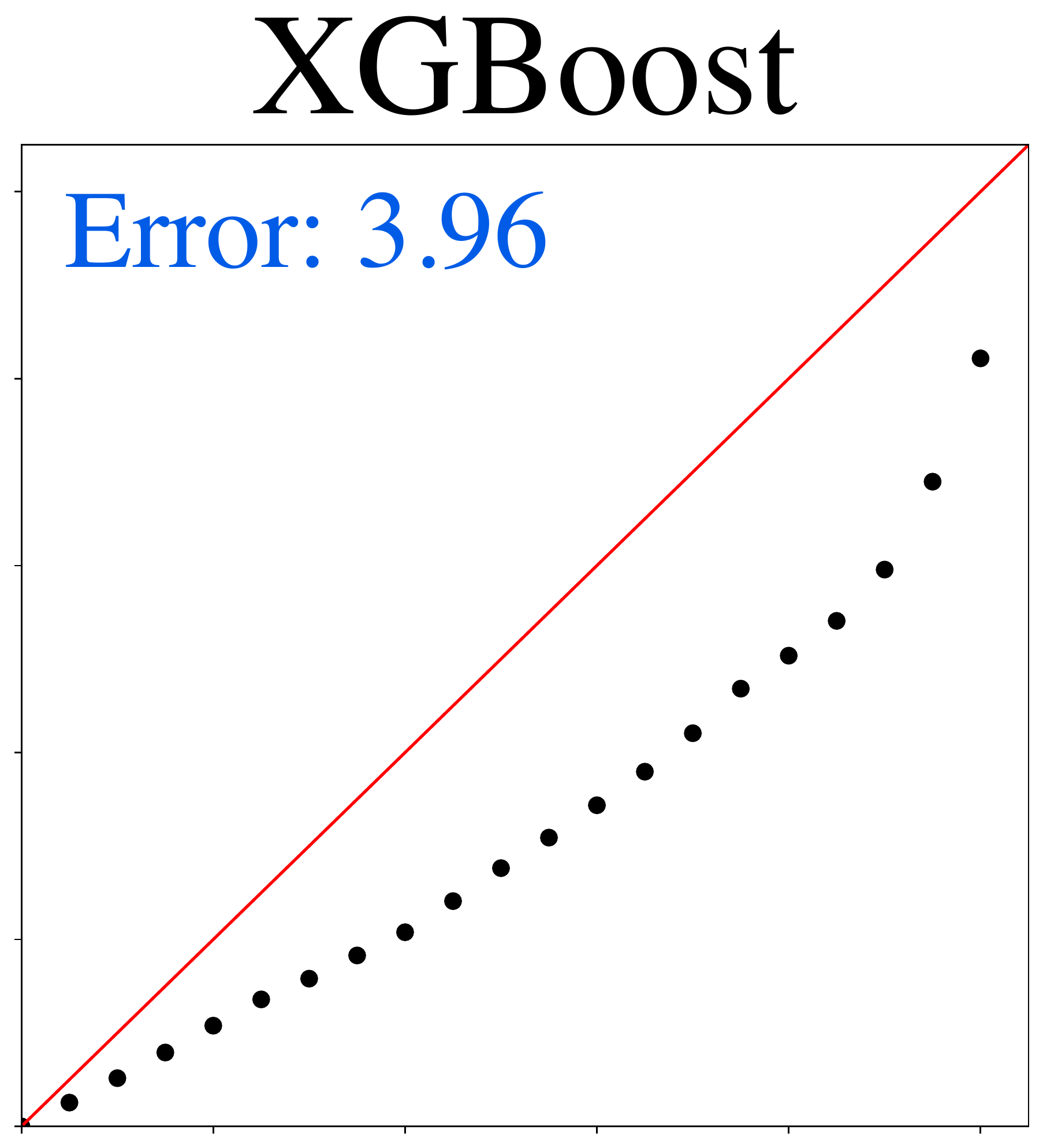}}  \\
 \multicolumn{5}{l}{\includegraphics[scale=0.093]{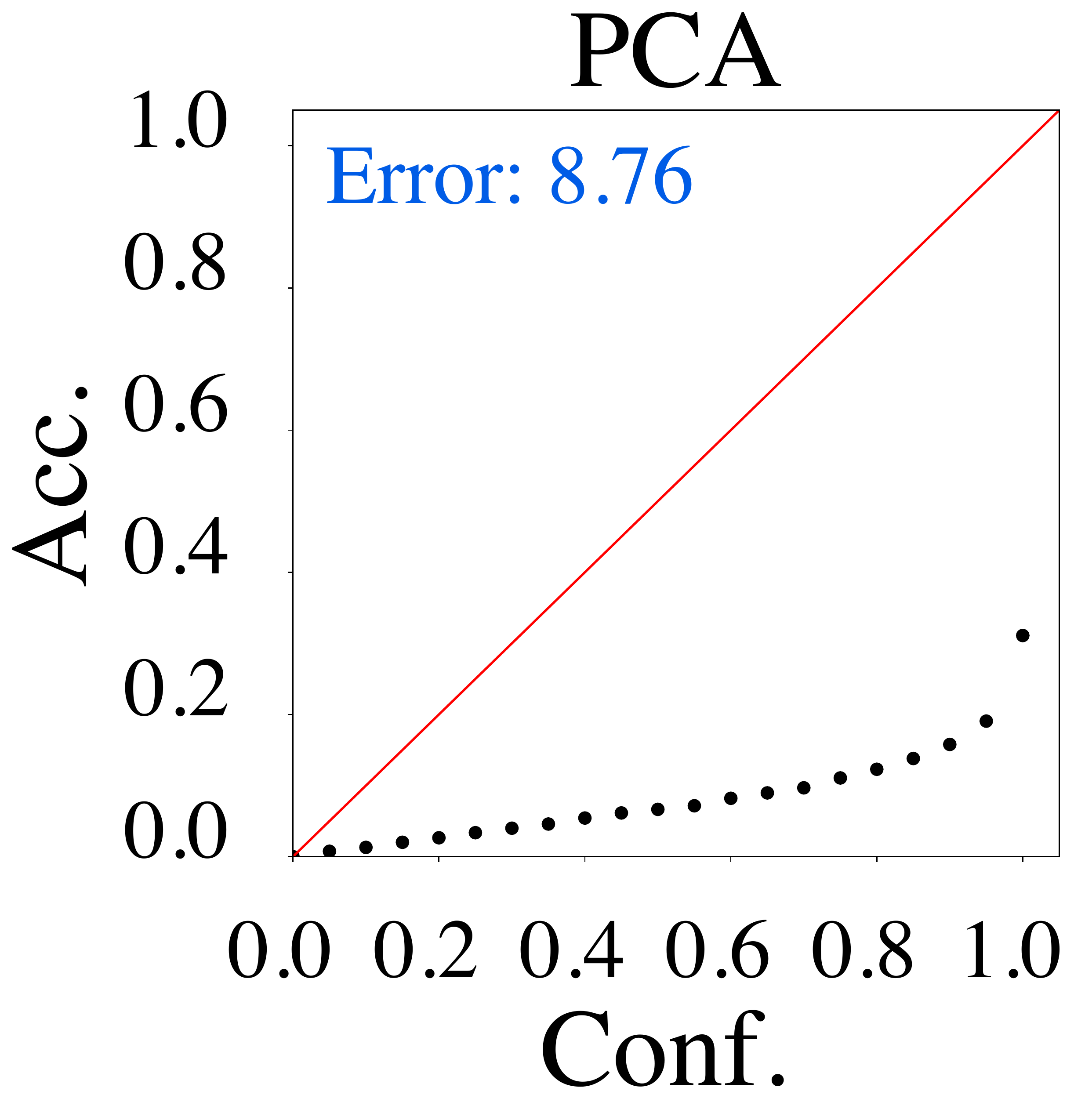}} &  \multicolumn{7}{l}{\includegraphics[scale=0.093]{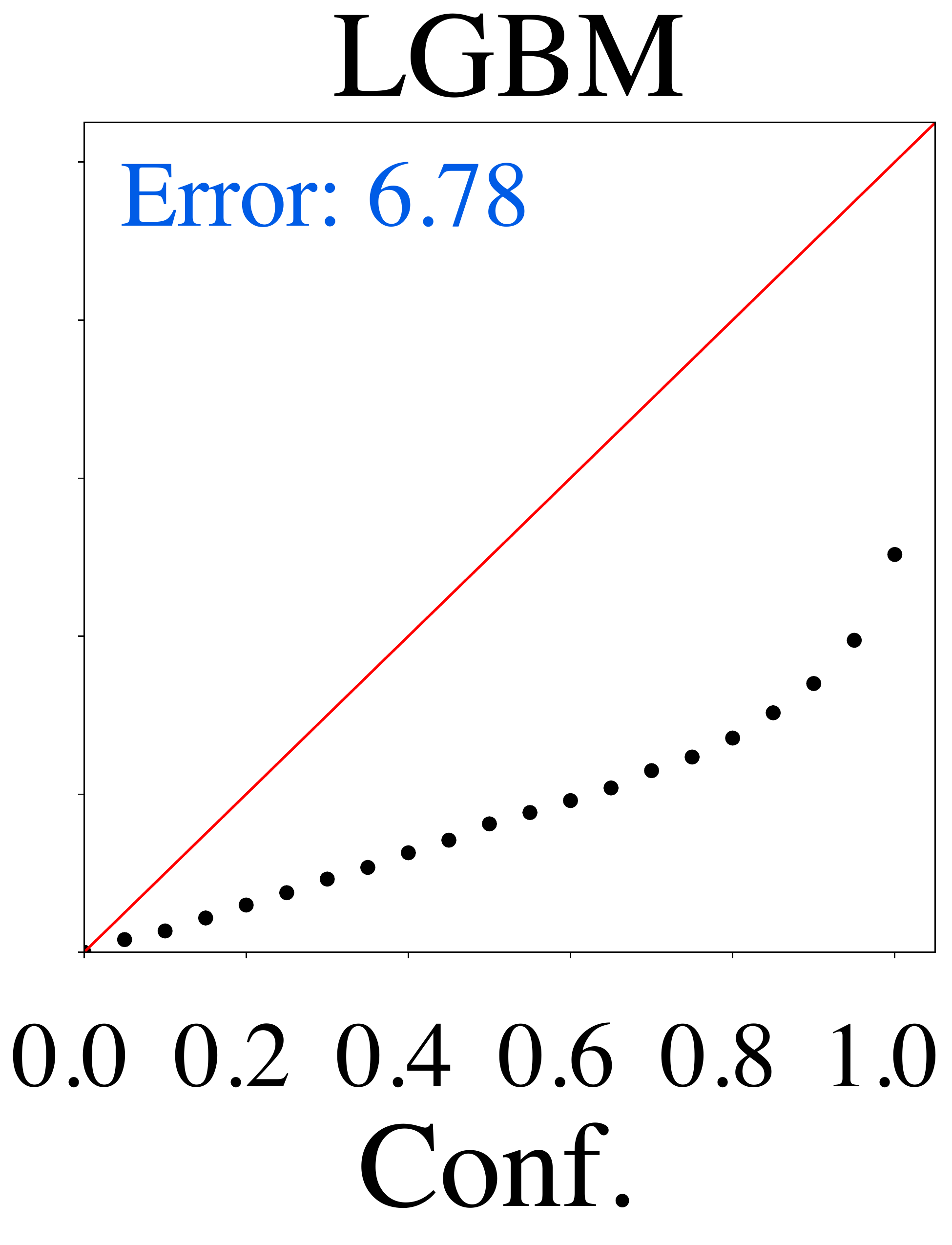}} & &&&&&& && \multicolumn{7}{l}{\includegraphics[scale=0.093]{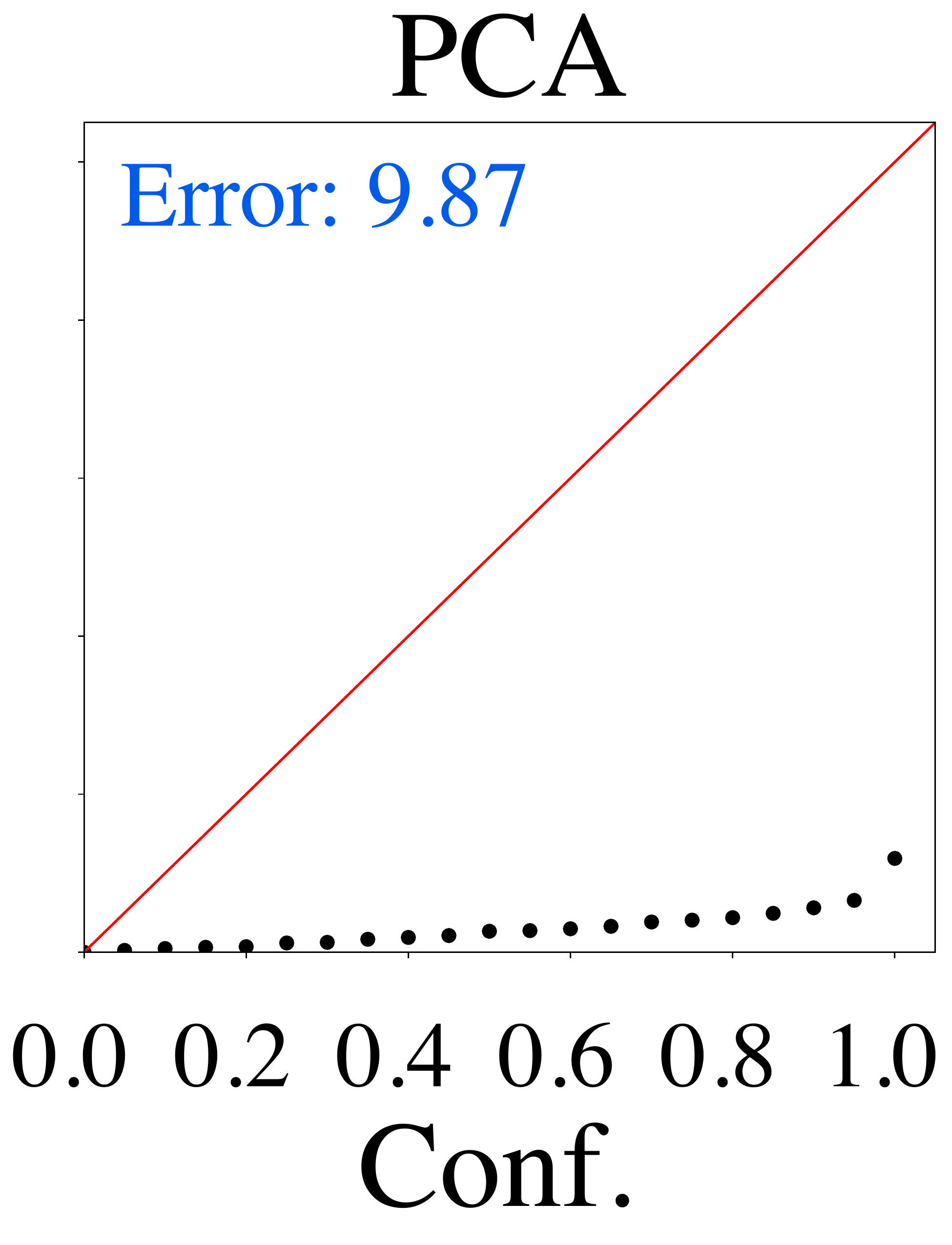}} & \multicolumn{7}{l}{\includegraphics[scale=0.093]{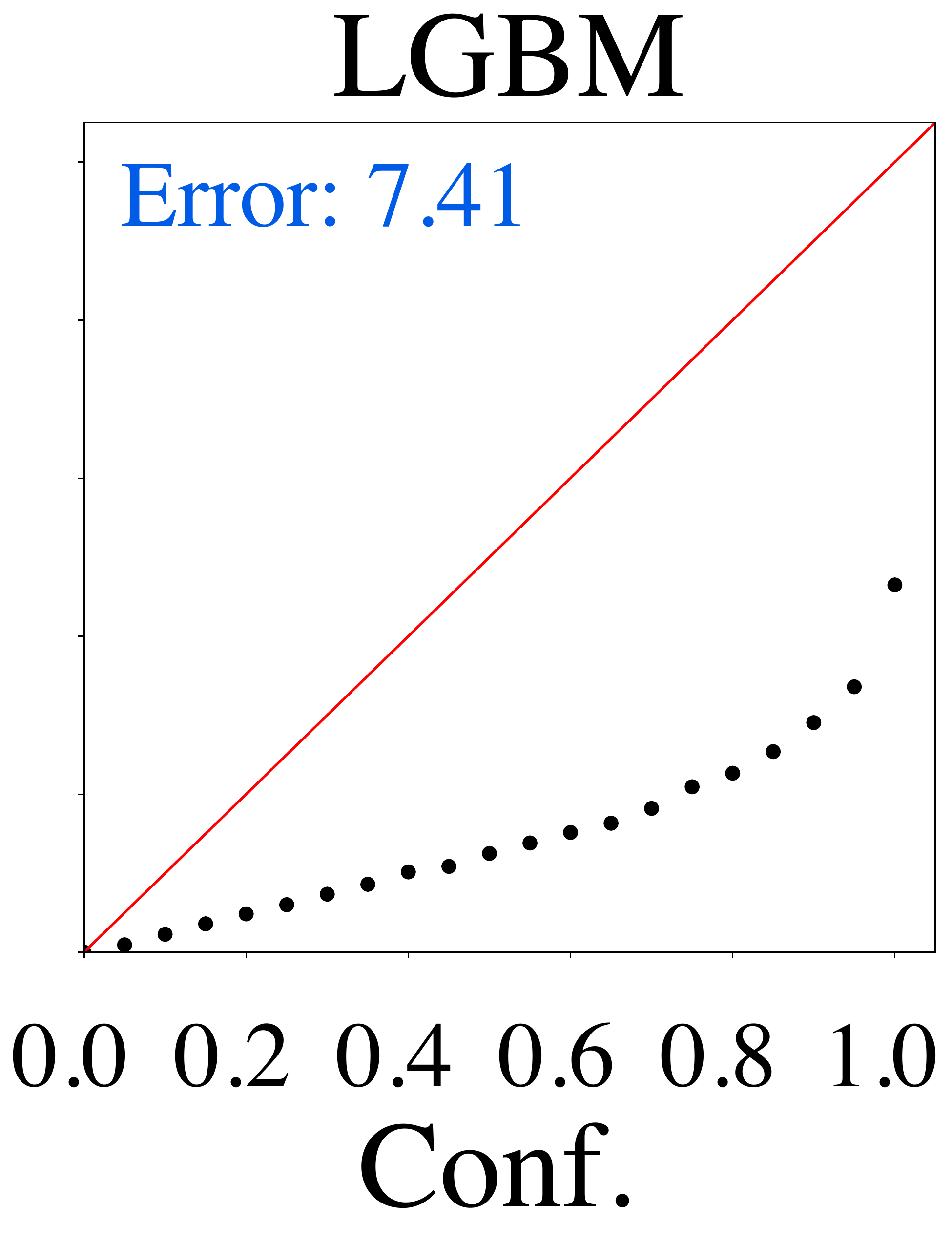}}&&&&&&&&& \multicolumn{7}{l}{\includegraphics[scale=0.093]{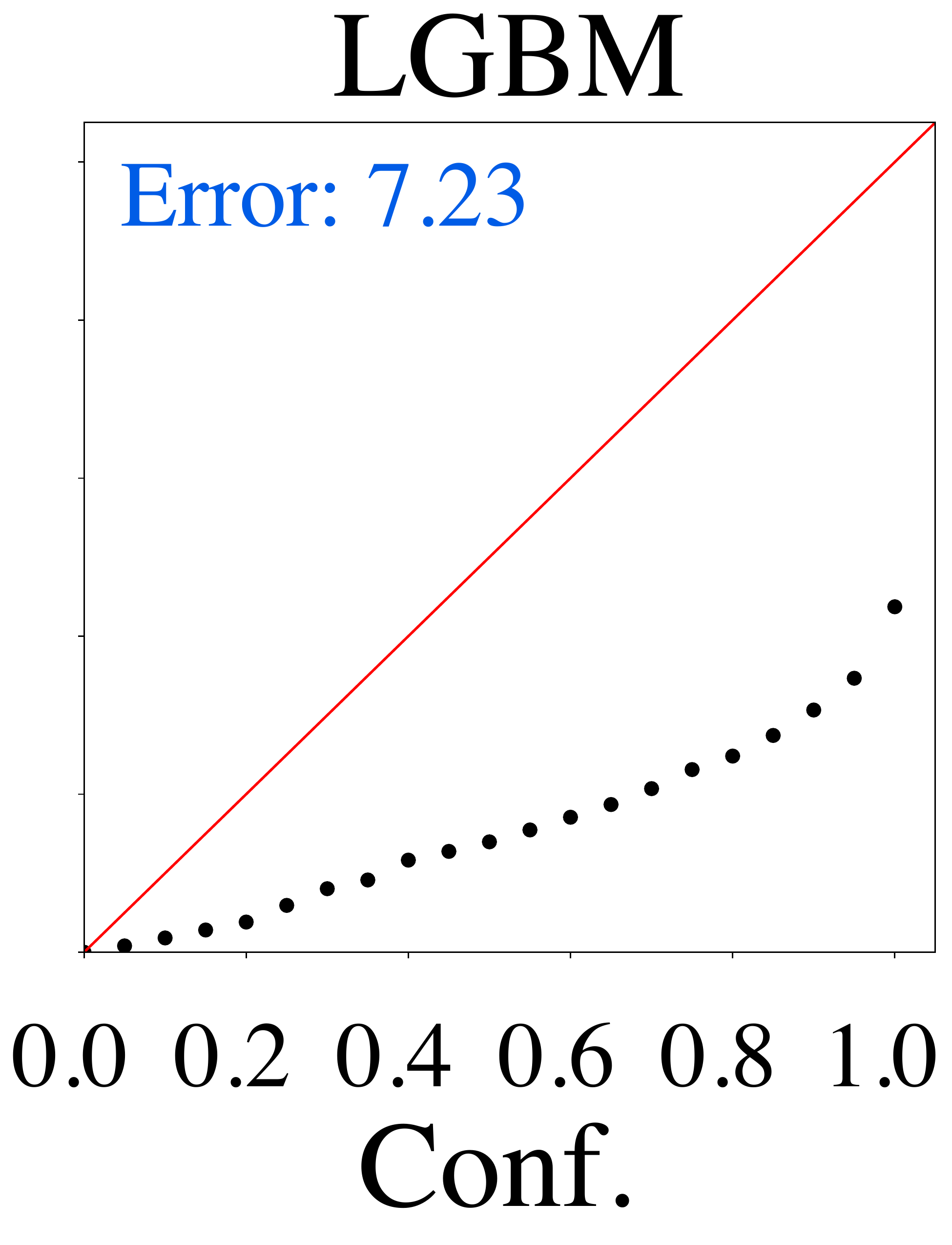}} &&&&&&&&&  \multicolumn{7}{l}{\includegraphics[scale=0.093]{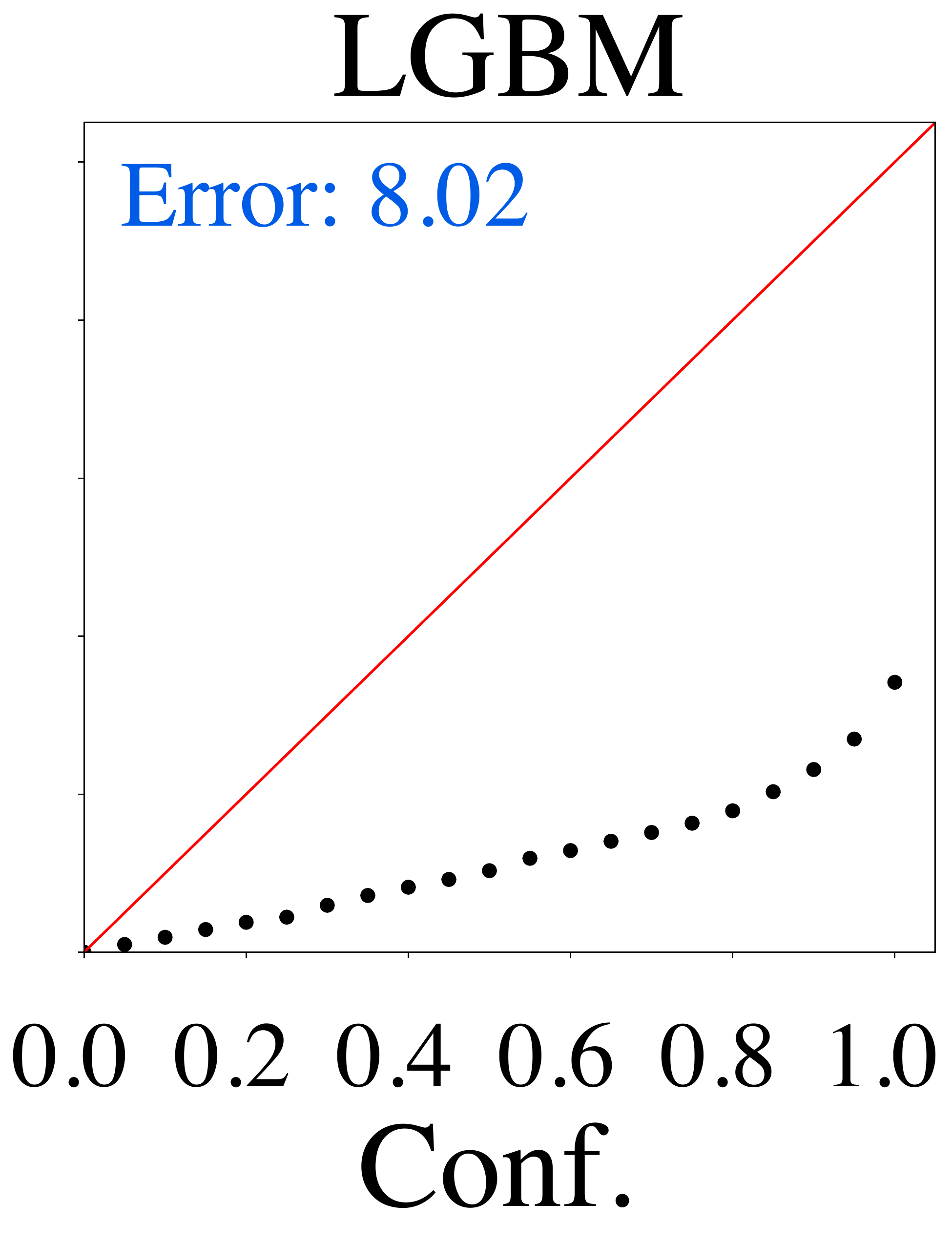}}  \\
    \multicolumn{13}{c}{(a) NER} & &&&&&& && \multicolumn{15}{c}{(b) CWS} &&&&&&&&& \multicolumn{7}{c}{(c) MT}    &&&&&&&&& \multicolumn{7}{c}{(d) POS}  \\
    \end{tabular}
    \vspace{-7pt}
      \caption{Calibration of different performance prediction models on four tasks (\texttt{NER}, \texttt{CWS} for fine-grained evaluation setting and \texttt{MT}, \texttt{POS} for holistic evaluation setting). 
      } 
      
  \label{fig:Reliability}%
\end{figure*}%

\paragraph{Setup}
As described in \S\ref{sec:ci}, we use non-parametric bootstrap to produce confidence intervals for $\hat{y}$. 
For the holistic evaluation setting, we do not include tensor-based models since the property, ``\textit{with replacement}'', of the bootstrap makes it difficult to construct resampled tensors in the holistic evaluation setting.

When calculating a calibration error as defined in Eq.~\ref{eq:CE}, we set $M=20$, choosing a range of 20 increasing confidence levels,  (i.e. $\gamma_1=0.05, \gamma_{2}=0.10 \cdots, \gamma_{20} = 1.00$), to evaluate the correctness
of the confidence intervals given by prediction models.
Besides using a reliability diagram and a calibration error, to compare the calibration performances of different models more comprehensively, we additionally use the following quantitative metrics: 
(1) \textit{average width} is the mean range of all the prediction distributions, formally, the difference between the maximum and minimum: 
$\frac{1}{N}\sum_{i\in[1,N]}
\max(\hat{Y}_i) - \min(\hat{Y}_i)$. 
(2) \textit{coverage} is the value of $\mathrm{acc}_b$ evaluated at $\gamma=1$. (i.e. proportion of ${y}$ $\hat{y}$ that fall into the distributions $\hat{Y}$, out of all the $N$ prediction entries).

\renewcommand\tabcolsep{3pt}

\begin{table}[!htb]
  \centering \small
    \begin{tabular}{lcccccc}
    \toprule
    \multirow{2}[4]{*}{\textbf{Model}} & \textbf{CE} & \textbf{Wid.} & \textbf{Cov.} & \textbf{CE} & \textbf{Wid.} & \textbf{Cov.} \\
\cmidrule(r){2-4} \cmidrule(r){5-7}          & \multicolumn{3}{c}{\textbf{NER}} & \multicolumn{3}{c}{\textbf{CWS}} \\
    \midrule
    Baseline & \cellcolor{mygrey}10.47  & \cellcolor{mygrey}0.006  & \cellcolor{mygrey}0.01  & 10.50  & 0.003  & 0.01  \\
    XGBoost & \cellcolor{mygrey}4.60  & \cellcolor{mygrey}0.093  & \cellcolor{mygrey}0.75  & 4.38  & 0.045  & 0.77  \\
    LGBM  & \cellcolor{mygrey}6.78  & \cellcolor{mygrey}0.093  & \cellcolor{mygrey}0.50  & 7.41  & 0.029  & 0.46  \\
    CP    & \cellcolor{mygrey}6.33  & \cellcolor{mygrey}0.110  & \cellcolor{mygrey}0.68  & 5.22  & 0.099  & 0.85  \\
    PCA   & \cellcolor{mygrey}8.76  & \cellcolor{mygrey}0.051  & \cellcolor{mygrey}0.31  & 9.87  & 0.015  & 0.12  \\
    \midrule
    \textbf{Model} & \multicolumn{3}{c}{\textbf{MT}} & \multicolumn{3}{c}{\textbf{POS}} \\
    \midrule
    Baseline & \cellcolor{mygrey}9.98  & \cellcolor{mygrey}1.46  & \cellcolor{mygrey}0.08  & 10.30  & 3.96  & 0.03  \\
    XGBoost  & \cellcolor{mygrey}3.75  & \cellcolor{mygrey}5.55  & \cellcolor{mygrey}0.81 & 3.96  & 17.61  & 0.82  \\
    LGBM    & \cellcolor{mygrey}7.23  & \cellcolor{mygrey}3.01  & \cellcolor{mygrey}0.44  & 8.02  & 12.90  & 0.34 \\
    \bottomrule
    \end{tabular}%
    \vspace{-7pt}
    \caption{Calibration errors (CE), average width (Wid.), coverage (Cov.) of different models over four tasks. (\texttt{NER}, \texttt{CWS} for fine-grained evaluation setting and \texttt{MT}, \texttt{POS} for holistic evaluation setting)}
\label{tab:metric-calibration}%
\end{table}%

\paragraph{Results}
The reliability diagram of different models and their corresponding metrics on four tasks are illustrated in Fig.~\ref{fig:Reliability} and Tab.~\ref{tab:metric-calibration} respectively.
Intuitively, the smaller the CE (calibration error) value is, the more closely the black dotted line becomes diagonal. Ideally, a perfectly calibrated model should have a CE of 0.

From these two tables, we see that:
(1) Overall, in both holistic (\texttt{MT} and \texttt{POS})  and fine-grained settings (\texttt{NER} and \texttt{CWS}), we see that XGBoost achieves the lowest calibration error together with a higher coverage, especially in the holistic setting.
(2) We observe that all of the plots indicate that the intervals produced by the models are over-confident, as the dots lie under the identity function. In other words, given a confidence level $\gamma$, the actual accuracy is lower than $\gamma$.
(3) In Tab.~\ref{tab:result-RMSE}, we find that LGBM achieves the lowest RMSE (2.389) in task MT, but its calibration error (7.23) is worse than XGBoost (3.75), implying that a model that predicts accurately is not necessarily well calibrated. This could be explained by the observation that the predicted distribution $\hat{Y}$ of LGBM has a narrower width (3.01). Given a large number of trials predicted by LGBM, we cannot be confident that the true $y$ is contained in the range of values predicted. 

\begin{figure}[htb]
    \centering 
     \subfloat[Bias]{
    \includegraphics[width=0.4\linewidth]{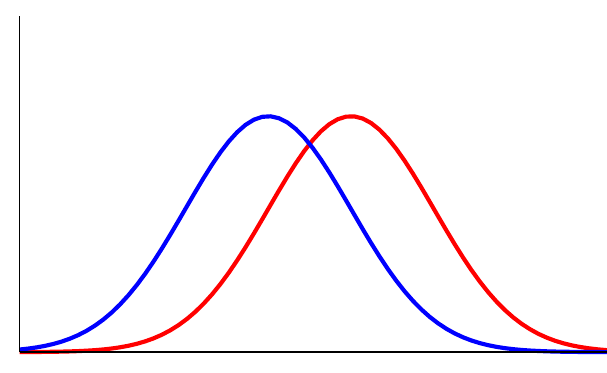}
    }    \hspace{-0.1em}
    \subfloat[Over-Confident]{
    \includegraphics[width=0.44\linewidth]{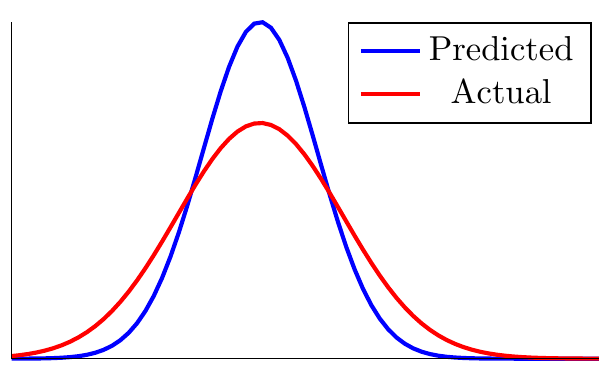}
    } \hspace{-0.1em} \\
    \subfloat[Bias]{
    \includegraphics[width=0.47\linewidth]{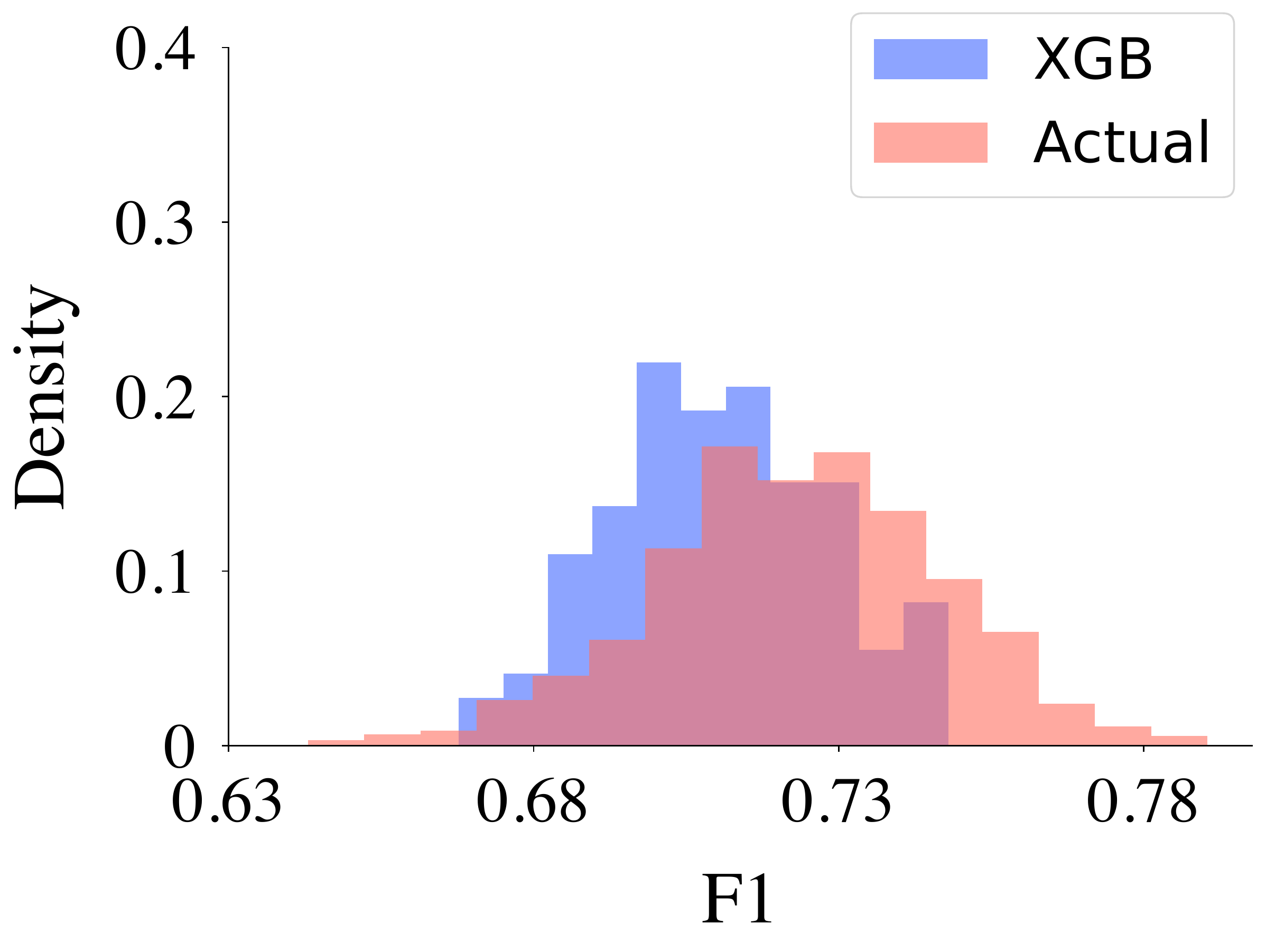}
    }
    \subfloat[Over-Confident]{
    \includegraphics[width=0.47\linewidth]{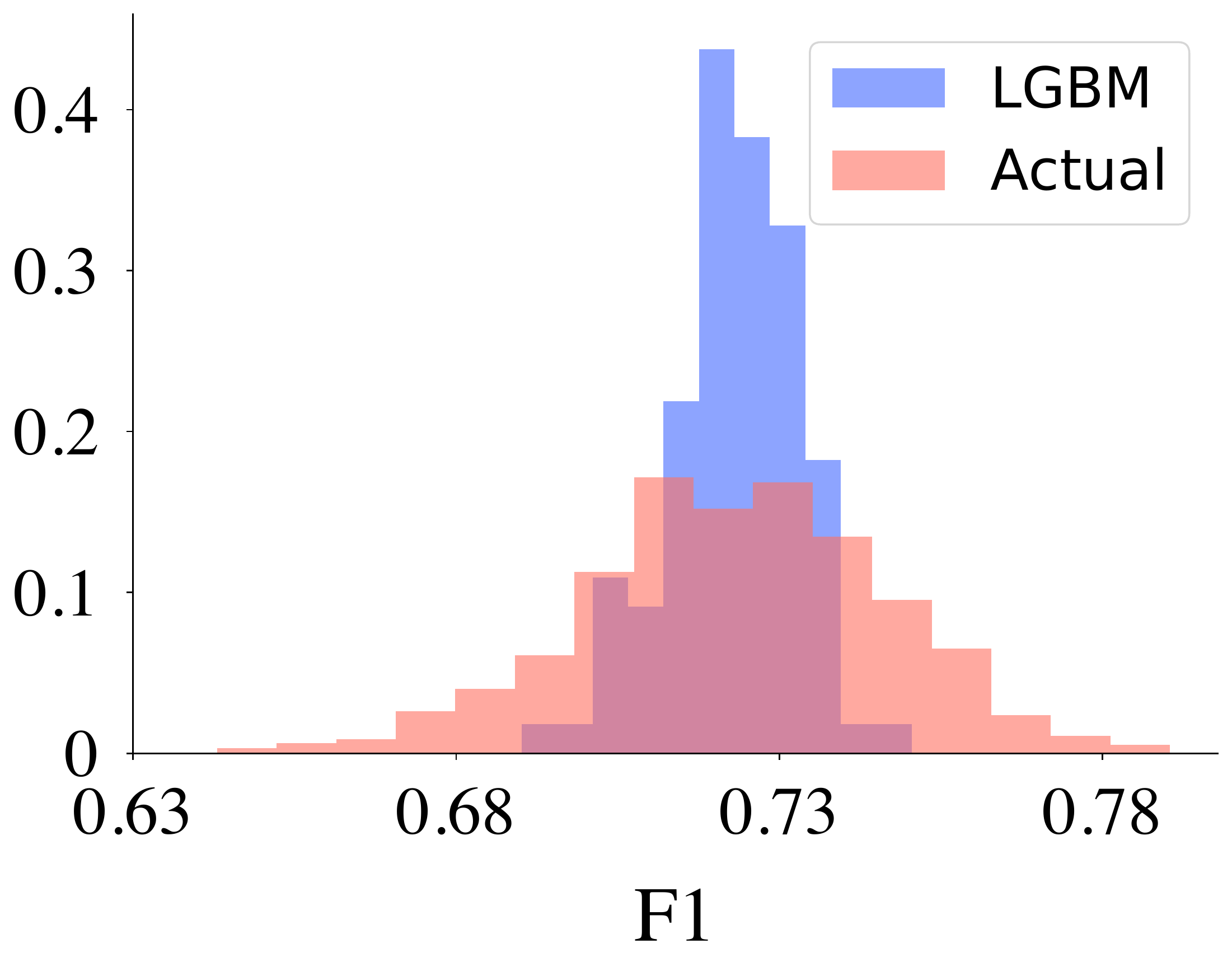}
    }\hspace{-0.1em}
    \vspace{-7pt}
    \caption{The first row of two plots (a,b)  artificially simulate two typical relations between actual and predicted distributions. The second row of two plots (c,d) show two real-world distributions of predicted performance w.r.t one test sample from NER task against corresponding actual distributions.} 
    \label{fig:case-study}
\end{figure}

\paragraph{Case Analysis}
To get a better understanding of how calibration analysis is conducted on different performance models, we perform a case study on \texttt{NER} task.
Fig.~\ref{fig:case-study}(a-b) illustrates two plots that artificially simulate two common relations following \citet{Diebold1997} between actual and predicted distribution: (i)  \textit{Bias} (ii) \textit{Over-confidence}.
From Tab.~\ref{tab:metric-calibration}, we see that XGBoost is better calibrated than LGBM in NER task. To interpret this gap, we (i) first randomly select test samples from NER dataset and then (ii) use two performance prediction models {XGBoost} and {LGBM} to produce blue distributions in Fig.~\ref{fig:case-study}(c-d) using the bootstrap (as \S\ref{sec:ci}). A perfectly calibrated model will show a histogram shape that resembles the actual one. We can see that the histogram shape of (d) signifies an over-confidence problem, in which the predicted distribution (in blue) is covered by the actual distribution (in red).
By contrast, in (c) the histogram of XGBoost in blue shifts to the left compared with the actual observed distribution, indicating that the prediction on this bucket is biased.



\section{Implications and Future Directions}
In this work, we not only widen the applicability of performance prediction, extending it to fine-grained evaluation scenarios, but also establish a set of reliability analysis mechanisms to improve its practicality.
In closing, we highlight some potential future directions:

\noindent\textbf{Confidence over confidence}:
Our work provides an idea for reliability analysis of the predicted confidence interval, which could also be explored on other scenarios, e.g., density forecasting \cite{Diebold1997}.
Another potentially valuable research topic is to build connections with the probability integral transform \cite{angus1994probability}, which is a typical method of calibration evaluation in financial risks, and our proposed calibration method.

\noindent\textbf{Calibration for automated evaluation metrics}:
From a broader point of view,  the role of existing learnable automatic evaluation metrics for text generation, such as BLEURT \cite{sellam-etal-2020-bleurt} and COMET \cite{rei2020comet}, is similar to a performance prediction model (i.e.,~both take features of input data as input and then output an evaluation score). Reliability analysis of these metrics is also an important topic since they determine the direction of model optimization.

\section*{Acknowledgements}
We sincerely thank all reviewers for their insightful comments and suggestions. We also thank Mengzhou Xia for discussions on details of performance prediction for different NLP tasks. 
This work was supported in part by a grant under
the Northrop Grumman SOTERIA project and the
Air Force Research Laboratory under agreement
number FA8750-19-2-0200.
The U.S. Government
is authorized to reproduce and distribute reprints for Governmental
purposes notwithstanding any copyright notation thereon. The views and
conclusions contained herein are those of the authors and should not be
interpreted as necessarily representing the official policies or
endorsements, either expressed or implied, of the Air Force Research
Laboratory or the U.S. Government.

\bibliographystyle{acl_natbib}
\bibliography{tensor}

 \appendix

\section{Datasets}

\label{sec:datasets}
\paragraph{Named Entity Recognition (NER)} We choose two well-established benchmark datasets: CoNLL-2003~\footnote{https://www.clips.uantwerpen.be/conll2003/ner/} and OntoNotes 5.0. \footnote{https://catalog.ldc.upenn.edu/LDC2013T19} 
CoNLL-2003 is drawn from Reuters news.
OntoNotes 5.0 is collected from newsgroups (\texttt{NW}), broadcast news (\texttt{BN}), broadcast conversation (\texttt{BC}), weblogs (\texttt{WB}), magazine genre (\texttt{MZ}), and telephone speech (\texttt{TC}), in which the first five genres of text are used in this paper.

\paragraph{Chinese Word Segmentation (CWS)} We consider five mainstream datasets: \texttt{CKIP}, \texttt{CTB}, \texttt{MSR}, \texttt{NCC}, and \texttt{SXU}, from SIGHAN2005~\footnote{http://sighan.cs.uchicago.edu/bakeoff2005/} and SIGHAN2008~\footnote{https://www.aclweb.org/mirror/ijcnlp08/sighan6/chinesebakeoff.htm}. The traditional Chinese characters in \texttt{CKIP} are mapped to simplified Chinese characters in our experiment.

\section{Models}
Our NER (CWS) models can be decomposed into four aspects: 1) character/subword encoders; 2) word (bigram) embeddings; 3) sentence-level encoders; 4) decoders. 

The four aspects of NER can be summarized as:
1) character/subword encoder: \texttt{ELMo} \citep{peters2018deep}, \texttt{Flair} \citep{akbik2018contextual,akbikpooled}, \texttt{BERT}~\footnote{BERT is grouped into the subword encoder because we use it to obtain the representation of subwords.} \citep{peters2018deep,devlin2018bert}; 
2) additional word embeddings: \texttt{GloVe} \citep{pennington2014glove};
3) sentence-level encoders: \texttt{LSTM} \citep{hochreiter1997long}, \texttt{CNN} \citep{kalchbrenner2014convolutional,chen2019grn:};
4) decoders: \texttt{MLP} or \texttt{CRF} \citep{lample2016neural,collobert2011natural}.  

The four aspects' setting of CWS: 1) character/subword encoder: \texttt{ELMo}, \texttt{BERT}; 2) bigram embeddings: \texttt{Word2Vec} \citep{mikolov2013efficient}, averaging the embedding of two contiguous characters; the settings of 3) the sentence-level encoders and 4) decoders are equal to NER.

We can also do bootstrap for predictions using regression models. If we consider recovering missing data with CP decomposition as a prediction method, we can construct a CI on the predicted values too.

\section{Hyper-parameters}
For XGBoost, we use squared error as the objective function for regression and set the learning rate as 0.1. We allow the maximum tree depth to be 10, the number of trees to be 100, and use the default regularization terms to prevent the model from overfitting.
For LGBM, we set the objective as regression for LGBMRegressor, the number of boosted trees and maximum tree leaves to be 100, adopt a learning rate of 0.1, and use the default regularization terms. For the Robust PCA model, we scale all the datasets, adopt the default regularization parameter of 1 for both the low rank and the sparse tensor, and set the learning rate as 1.1.
For CP Decomposition, we do not standardize the features in CWS and NER, but do so for WMT and POS. We adopt a rank $r=5$ in training and performance prediction, expressing the recovered tensor used for prediction to be a sum of 5 rank-1 tensors. 

\paragraph{Statistics of Tensor}
Tab.~\ref{tab:stat}, where \textit{sparsity} denotes the percentage of missing values in the tensor. 
\renewcommand\tabcolsep{3pt}
\begin{table}[t]
  \centering \footnotesize
    \begin{tabular}{lcccc}
    \toprule
    \textbf{Feature} & \textbf{NER} & \textbf{CWS} & \textbf{MT} & \textbf{POS} \\
    \midrule
    Sparisity  & 0.0   & 0.0   & 0.346  & 0.019  \\
    Shape & (11,6,9,4) & (5,8,8,4) & (39,39,22) & (26,60,14) \\
    \bottomrule
    \end{tabular}%
    \caption{Statistics of performance tensors for four tasks. \textit{Sparsity} denotes the percentage of missing values in the tensor}
      \label{tab:stat}%
\end{table}%

\end{document}